\documentclass[11pt]{article}
\usepackage{amsmath}
\usepackage{amssymb}
\usepackage{amsfonts}

\usepackage[preprint]{acl} 

\usepackage{times}
\usepackage{latexsym}
\usepackage{booktabs}
\usepackage{multirow}
\usepackage{colortbl}
\usepackage[table]{xcolor}
\usepackage{tcolorbox}
\definecolor{ourblue}{RGB}{235, 242, 250}
\usepackage{tabularx}

\usepackage[T1]{fontenc}
\usepackage[utf8]{inputenc}
\usepackage{microtype}
\usepackage{inconsolata}
\usepackage{graphicx}
\usepackage{enumerate}
\usepackage{enumitem}
\usepackage{makecell}

\title{Interleaved Latent Visual Reasoning with Selective Perceptual Modeling}

\author{
    \textbf{Shuai Dong}\textsuperscript{1,2}, \quad 
    \textbf{Siyuan Wang}\textsuperscript{3}\thanks{\ \ Corresponding authors.}, \quad 
    \textbf{Xingyu Liu}\textsuperscript{1}, \\
    \textbf{Chenglin Li}\textsuperscript{2,5}, \quad 
    \textbf{Haowen Hou}\textsuperscript{2,6}, \quad 
    \textbf{Zhongyu Wei}\textsuperscript{2,4}\footnotemark[1] \\
    \textsuperscript{1}China University of Geosciences, Wuhan \quad \textsuperscript{2}Shanghai Innovation Institute \\
    \textsuperscript{3}University of Southern California \quad \textsuperscript{4}Fudan University \\
    \textsuperscript{5}Zhejiang University \quad \textsuperscript{6}Shanghai Jiao Tong University \\
    \texttt{\{dongshuai\_iu, liuxingyu\}@cug.edu.cn, sw\_641@usc.edu} \\
    \texttt{22351307@zju.edu.cn, haowenhou@outlook.com, zywei@fudan.edu.cn}
}

\begin{document}
\maketitle
\begin{abstract}
Interleaved reasoning paradigms enhance Multimodal Large Language Models (MLLMs) with visual feedback but are hindered by the prohibitive computational cost of re-encoding pixel-dense images. A promising alternative, latent visual reasoning, circumvents this bottleneck yet faces limitations: methods either fail to capture intermediate state evolution due to single-step, non-interleaved structures, or sacrifice precise perceptual modeling by over-compressing features. We introduce Interleaved Latent Visual Reasoning (ILVR), a framework that unifies dynamic state evolution with precise perceptual modeling. ILVR interleaves textual generation with latent visual representations that act as specific, evolving cues for subsequent reasoning. Specifically, we employ a self-supervision strategy where a momentum teacher model selectively distills relevant features from ground-truth intermediate images into sparse supervision targets. This adaptive selection mechanism guides the model to autonomously generate context-aware visual signals. Extensive experiments on multimodal reasoning benchmarks demonstrate that ILVR outperforms existing approaches, effectively bridging the gap between fine-grained perception and sequential multimodal reasoning. The code is available at \url{https://github.com/XD111ds/ILVR}.
\end{abstract}

\section{Introduction}
Multimodal Large Language Models (MLLMs)~\citep{Li2024LLaVAOneVisionEV,Bai2025Qwen25VLTR,Wang2025InternVL35AO} have demonstrated remarkable capabilities in bridging the gap between vision and language. Capitalizing on the reasoning prowess of Large Language Models (LLMs), recent works have successfully adapted Chain-of-Thought (CoT) methodologies to the multimodal domain~\citep{Zhang2023MultimodalCR,Bai2025UniVGR1RG,Huang2025VisionR1IR,Wei2022ChainOT}. This enables models to decompose complex visual tasks into sequential intermediate steps, achieving sophisticated reasoning grounded in visual content.

Recent work explores interleaved image-text reasoning by injecting intermediate visual images within textual CoTs to enhance multimodal understanding and planning~\citep{Shao2024VisualCU}. These approaches generally fall into two paradigms. The first uses external tools to statically manipulate the input image, e.g., highlighting key regions ~\citep{Fu2025ReFocusVE}, 
drawing auxiliary lines~\citep{Hu2024VisualSS}, or shifting image styles~\citep{Liu2025VisualAT}, to improve fine-grained perception. While relying on a single visual state, it cannot model evolving scenarios or simulate action outcomes crucial for sequential tasks~\citep{Li2025ZebraCoTAD}. The second paradigm addresses this employing a unified model to dynamically visualizing imagined intermediate or future states~\citep{Chern2024ANOLEAO,Deng2025EmergingPI}. 
However, integrating visual generation and reasoning into a unified model often degrades reasoning performance. More critically, both paradigms incur high computational cost from iteratively re-encoding pixel-dense images, severely hindering multi-step reasoning.

\begin{figure*}[t]
    \centering
    \includegraphics[width=0.85\linewidth]{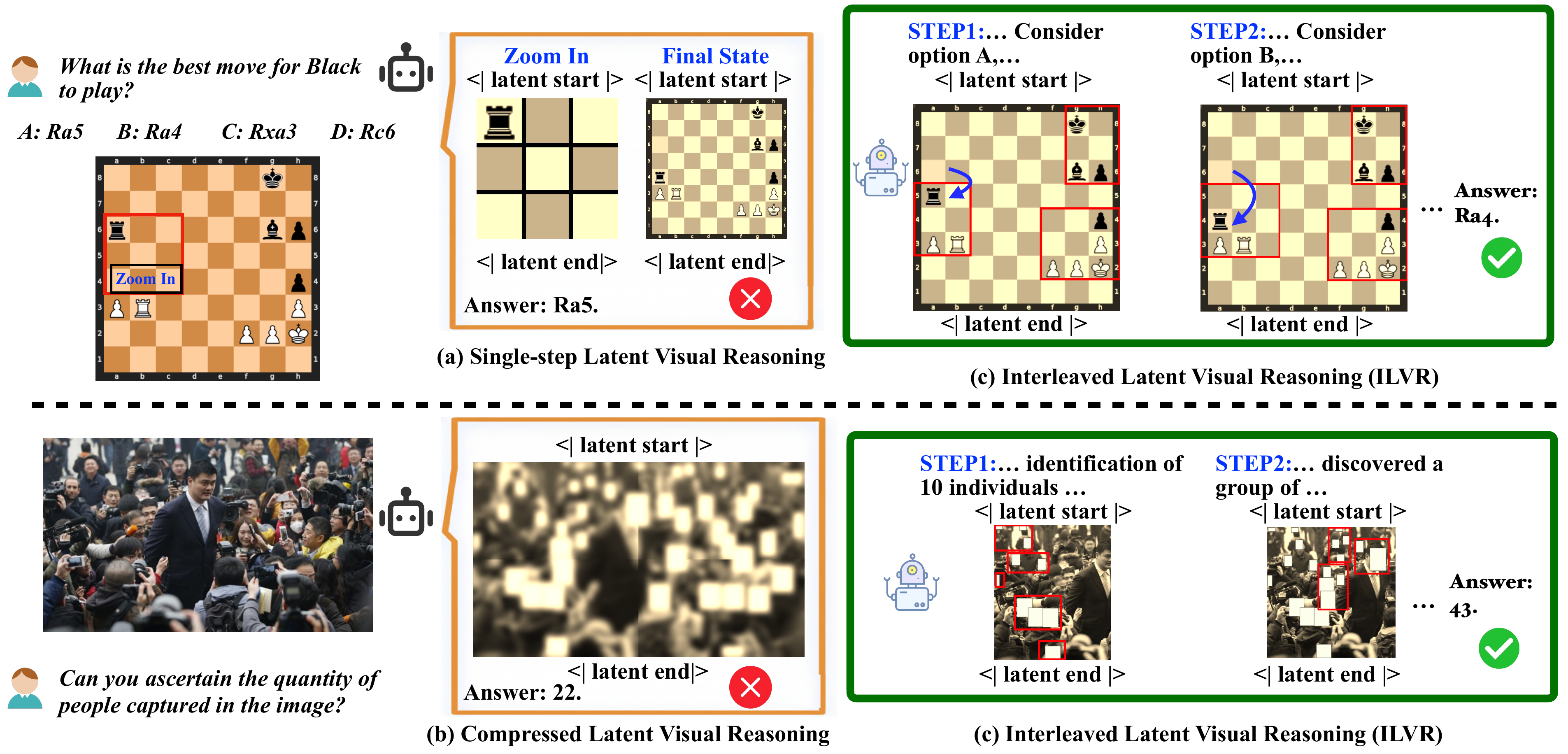}
    \caption{\textbf{Comparison of ILVR with prior latent visual reasoning methods.} 
    In the chess puzzle (top row), single-step approaches (a) either capture static initial details (e.g., a zoomed-in rook) or jump to a predicted final state, failing to model the hypothetical states needed to evaluate move options.
    In the dense counting task (bottom row), methods relying on heavily compressed latent representations (b) lose fine-grained details, resulting in a hallucinated count. 
    In contrast, our ILVR (c) succeeds by interleaving textual reasoning with dynamically updated latent states. Each latent representation provides essential visual cues for subsequent reasoning steps (highlighted in red boxes), unifying dynamic state evolution with precise perceptual modeling to reach the correct answer.}
    \label{fig:overview}
\end{figure*} 

Inspired by latent reasoning in LLMs~\citep{Shen2025CODICC,Hao2024TrainingLL}, the latent visual reasoning paradigm replaces explicit images with latent representations to avoid costly pixel-level processing. However, current methods face two major limitations. First, most adopt a single-step, non-interleaved design. For instance, LVR~\citep{Li2025LatentVR} and Mirage~\citep{Yang2025MachineMI} generate latent representations only once, either for a region of the static input image or the final state after all actions, and cannot model intermediate and evolving states during reasoning. In the chess puzzle in Fig.~\ref{fig:overview}(a), relying on a static zoom-in or a predicted final state is insufficient, as it bypasses step-by-step verification of move legality (e.g., path obstructions), often leading to erroneous predictions. Second, methods like Mirage~\citep{Yang2025MachineMI} derive latent representations by heavily compressing dense visual features from the entire image into limited latent tokens. As the counting task shown in Fig.~\ref{fig:overview}(b), such over-compression discards crucial perceptual details and leads to hallucination.

To this end, we propose Interleaved Latent Visual Reasoning (ILVR) framework to integrate dynamic latent visual reasoning with selective perceptual modeling. ILVR interleaves reasoning between explicit textual generation and latent visual representations that are continuously updated to capture the most relevant visual cues at each reasoning step.
We train the model to learn this interleaved paradigm by approximating ground-truth interleaved image-text trajectories, with textual outputs supervised using cross-entropy loss while latent representations are aligned with selectively extracted features from their corresponding images, which we refer to as \textit{helper images}. 
Specifically, we employ a momentum teacher model~\citep{He2019MomentumCF}, a temporally smoothed copy of the trained model, to selectively extract the most relevant features from \textit{helper images} by aggregating highly attended patches conditioned on the ongoing reasoning process. By internalizing this capability, ILVR effectively unifies precise perceptual modeling with dynamic evolution of latent visual states.

In summary, our contributions are threefold:
\begin{itemize}[leftmargin=13pt]
\setlength{\itemsep}{1pt}
\setlength{\parsep}{1pt}
\setlength{\parskip}{1pt}
    \item We propose Interleaved Latent Visual Reasoning (ILVR), a framework that interleaves explicit token generation with updated latent visual representations, enabling dynamic state evolution.
    \item We introduce an adaptive selection mechanism that distills the most relevant visual signals from the \textit{helper image} into latent representations at every reasoning step, using a self-supervised strategy guided by a momentum teacher model without requiring external supervision.
    \item Through extensive experiments on fine-grained visual perception and sequential planning tasks, we demonstrate ILVR's robust generalization in both in-domain and out-of-distribution (OOD) settings. 
    By operating entirely in latent space, it achieves up to 18× inference speedup over methods requiring costly explicit image generation.
\end{itemize}

\section{Related Work}
\subsection{Interleaved Image-Text Reasoning}
Interleaved image-text reasoning refers to the capability of models to generate intermediate visual feedback~\citep{Chern2024ANOLEAO,Li2025ZebraCoTAD,Deng2025EmergingPI}, either directly or via external tools~\citep{Hu2024VisualSS,Shao2024VisualCU,Su2025PixelRI}, to enhance their reasoning abilities. Early methods used external tools for static image edits, such as cropping or OCR \citep{Huang2025VisualToolAgentA, Zhang2025CMMCoTEC,Wang2025VGRVG}, but struggled to model evolving visual states. Recent generative approaches enable models to synthesize intermediate-state images \citep{Chern2024ANOLEAO,Deng2025EmergingPI}, yet they often face a trade-off between generative fidelity and reasoning performance. Crucially, both tool-based and generative paradigms suffer from high computational overhead due to repeated pixel-level encoding of dense visual data.

\subsection{Latent Reasoning}

To bypass discrete token constraints, latent reasoning performs multi-step inference in continuous hidden space~\citep{Shen2025CODICC,Hao2024TrainingLL,Cheng2024CompressedCO}. In the multimodal domain, Mirage~\citep{Yang2025MachineMI} precedes textual reasoning with a latent representation formed by encoding a problem-specific helper image and aggressively pooling its patch embeddings into highly compressed vectors. LVR~\citep{Li2025LatentVR} adopts a similar strategy but isolates key visual cues within a bounding box, generating latent representations of only that targeted region. Contemporaneous with our work, Sketchpad~\citep{Zhang2025LatentSS} also explores generating visual latents to elicit reasoning. However, a fundamental limitation plagues these approaches. In their paradigm, a model generates latent representations of a helper image once, and all subsequent steps are confined to pure textual reasoning. This non-interleaved structure inherently renders the visual information static and detached from the evolving reasoning trajectory.

\section{Method}
\begin{figure*}[t]
    \centering
    \includegraphics[width=0.9\linewidth]{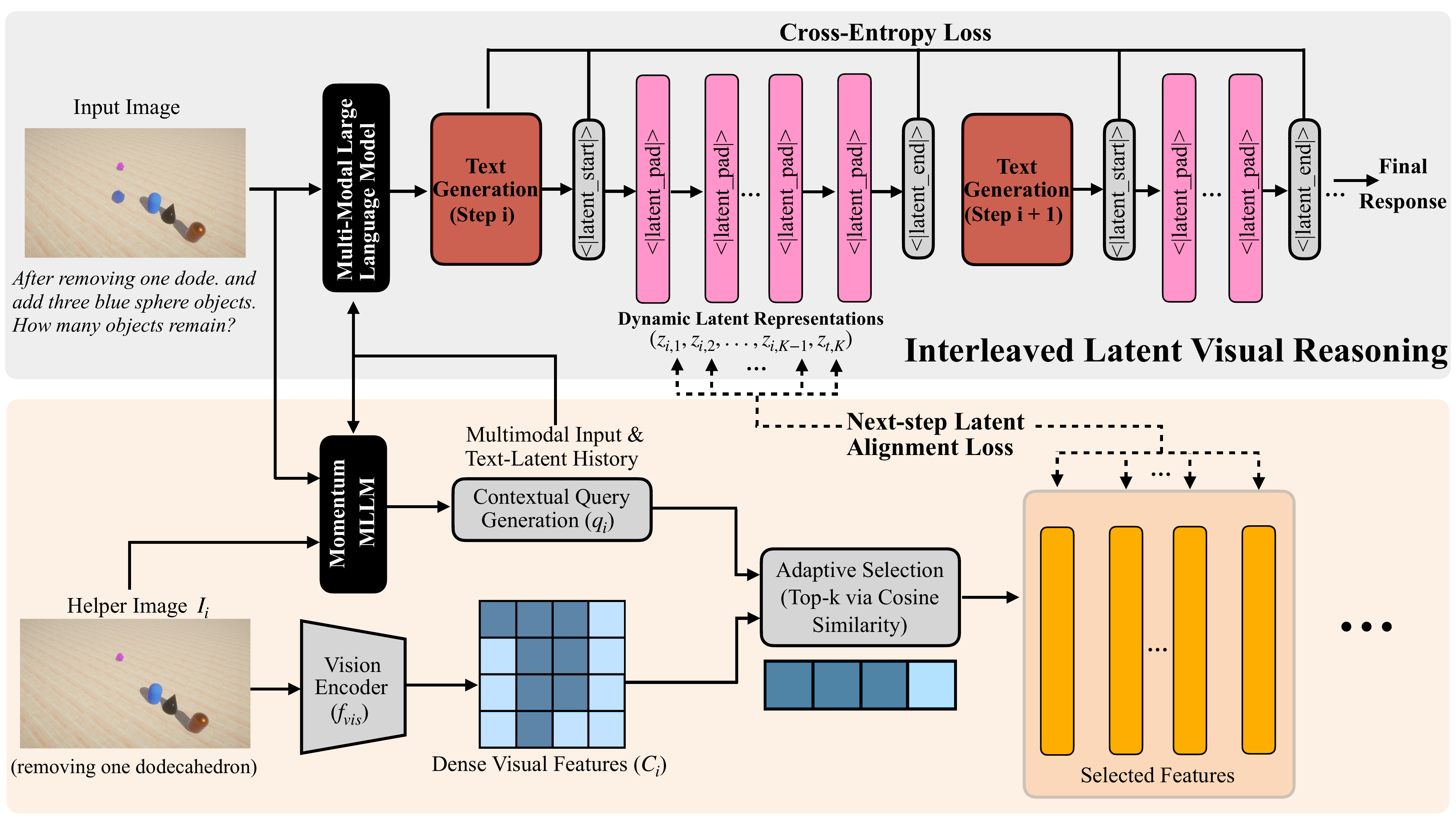}
    \caption{\textbf{The Interleaved Latent Visual Reasoning (ILVR) framework.} 
    The model performs multi-step reasoning by interleaving textual generation with dynamically evolving latent visual representations. 
    The momentum teacher model (bottom) utilizes the multimodal inputs and the text-latent history up to reasoning step $i$ to form a contextual query ($q_i$), which selectively extracts the most relevant visual features (yellow blocks) from the \textit{helper image}. 
    Simultaneously, the trained model (top) generates a sequence of latent representations (pink blocks) interleaved with reasoning text. 
    These latents are supervised via a next-step latent alignment objective that encourages them to match the teacher-selected visual features.}
    
    \label{fig:framework}
\end{figure*}

In this section, we present Interleaved Latent Visual Reasoning (ILVR) framework that performs reasoning by interleaving explicit textual generation with latent visual representations, as shown in Fig.~\ref{fig:framework}. We first outline the interleaved generation paradigm (Sec.~\ref{paradigm}). We then detail how we construct latent supervision targets by selecting key features from intermediate images (``\textit{helper images}'') within ground-truth interleaved image-text trajectories using a momentum teacher model (Sec.~\ref{supervision}). Finally, we describe the two-stage training strategy to instill this interleaved latent reasoning ability (Sec.~\ref{training}).

\subsection{Interleaved Text-Latent Paradigm}
\label{paradigm}
Our framework operates in an interleaved reasoning paradigm where the model autoregressively generates both text tokens and latent visual representations. The reasoning process is structured as a unified sequence $\mathcal{S}$ that alternates between textual tokens and latent segments:
\begin{equation}
\label{eq:seq_structure}
\small
\begin{split} 
    \mathcal{S} = [ & t_{1,1}, \dots, t_{1,M}, \texttt{<|latent\_start|>}, \\
    & z_{1,1}, \dots, z_{1,K}, \texttt{<|latent\_end|>}, \\
    & t_{2,1}, \dots, t_{2,N}, \texttt{<|latent\_start|>}, \\
    & z_{2,1}, \dots, z_{2,K}, \texttt{<|latent\_end|>}, \dots ]
\end{split}
\end{equation}
where $t_{i,j}$ denotes discrete text tokens and $z_{i,k}$ represents continuous latent embeddings at reasoning step $i$. The special tokens \texttt{<|latent\_start|>} and \texttt{<|latent\_end|>} explicitly delimit the boundaries of latent visual reasoning phases.

During inference, the model generates text tokens as usual. When the model produces a \texttt{<|latent\_start|>} token, it switches to a latent generation mode for a fixed length $K$. In this mode, instead of projecting the hidden state to the vocabulary size to sample a discrete token, the hidden state from the previous timestep $\mathbf{h}_{t}$ is fed directly as the input embedding for the current timestep, effectively bypassing the discrete embedding lookup $\mathbf{e}_{t+1} = \mathbf{h}_{t}$. The sequence of $K$ hidden states produced in this loop constitutes the model's self-generated latent representations. After completing $K$ latent generation, the model generates \texttt{<|latent\_end|>} and resumes explicit textual reasoning, utilizing the accumulated latent information as context.

To train the model with this paradigm, we utilize pre-constructed interleaved trajectories formatted as ``\texttt{reasoning text}$\rightarrow$\texttt{helper image}$\rightarrow$\texttt{reasoning text}$\rightarrow$\texttt{helper image} $\dots$''. We convert each trajectory into a unified supervision sequence by replacing each \textit{helper image} $I_i$ at reasoning step $i$ with a latent segment: a \texttt{<|latent\_start|>} followed by $K$ \texttt{<|latent\_pad|>} tokens, and terminated by \texttt{<|latent\_end|>}. The \texttt{<|latent\_pad|>} act as placeholders for the critical visual signals extracted from $I_i$. Thus, the core of our method is to select which visual features from $I_i$ should serve as regression targets to supervise the hidden states generated at these pad positions.

\subsection{Interleaved Supervision Construction}
\label{supervision}
To enable the model to generate meaningful latent representations, we employ a teacher model to construct high-quality supervision targets for the latent segments. Given the same reasoning context as the model being trained, the teacher processes the \textit{helper image} $I_i$ at reasoning step $i$ and extracts the most relevant visual features as ground-truth latent supervision. Meanwhile, the textual parts are trained using standard explicit text supervision.

\paragraph{Momentum Teacher Model}
We adopt a self-supervised strategy where the teacher is a momentum model, a temporally smoothed version of the model being trained (the student model). This design keeps the supervision signal stable and well-aligned with the evolving representation space of the student model. The parameters of the momentum model $\theta_m$ are updated as an Exponential Moving Average (EMA) of the student parameters $\theta$ with a decay factor $\tau$: $\theta_m \leftarrow \tau\theta_m + (1-\tau)\theta$.

\paragraph{Candidate Visual Feature Generation} The goal of the momentum teacher model is to selectively distill the pixel-dense \textit{helper image} into a sparse set of $K$ feature vectors most relevant to the current reasoning step. The teacher first encodes a \textit{helper image} $I_i$ using its frozen vision encoder $f_{\text{vis}}$ to obtain a dense pool of patch features:
\begin{equation}
    \mathbf{C}_i = f_{\text{vis}}(I_i) = \{ \mathbf{c}_{i,j} \in \mathbb{R}^H \}_{j=1}^{P_i},
\end{equation}
where $H$ is the hidden dimension and $P_i$ is the number of patches. 

However, raw patch features often suffer from varying information density depending on the image resolution. In high-resolution images, individual patches may capture only local textures rather than semantic concepts. To address this, we introduce a spatial aggregation step to adapt the feature density. Specifically, we set a threshold $L$: if the number of raw patches $P_i \geq L$, we pool features over local spatial windows to form a refined candidate pool $\mathbf{C}'_i$; otherwise, we retain the original fine-grained features. Formally,
\begin{equation}
    \mathbf{C}'_i = 
    \begin{cases} 
    \text{GroupMean}(\mathbf{C}_i, L), & \text{if } P_i \geq L \\
    \mathbf{C}_i, & \text{if } P_i < L
    \end{cases}
\end{equation}
where $\text{GroupMean}$ aggregates the $P_i$ patch sequence into $L$ semantic units, ensuring that the subsequent selection operates on robust features regardless of input resolution.

\paragraph{Teacher-Guided Selective Perceptual Modeling} The teacher model then identifies the most relevant candidate features from $\mathbf{C}'_i$ as supervision. It constructs a context-aware query $\mathbf{q}_i$ using the same context as the student model, including the multimodal inputs and the reasoning history up to step $i$. By computing cosine similarity between $\mathbf{q}_i$ and each feature in $\mathbf{C}'_i$, the teacher selects the top-$K$ features to form the supervision set $\mathbf{Z}_i$.

To construct the query $\mathbf{q}_i$, we do not apply naive average pooling over the entire context that would weaken critical signals. Instead, we separately process the input text, input image, and reasoning history, with the first two forming a global intent vector and the last providing local reasoning context. 

For dense input text, we apply mean pooling over their final-layer hidden states to obtain $\mathbf{r}_{\text{txt}}$. For sparse input images, we compute text-guided attention over image to selectively emphasize informative regions, yielding $\mathbf{r}_{\text{img}}$. The global intent vector is obtained by averaging the representation of input text and image as $\mathbf{u}=\frac{1}{2}(\mathbf{r}_{\text{txt}}+\mathbf{r}_{\text{img}})$.

To capture evolving reasoning dynamics, we incorporate the reasoning history up to step $i$ by averaging the final-layer hidden states of all textual rationales from step $1$ to step $i$ as $\mathbf{q}^{\text{text}}_{[1,i]}$
and all latent rationales from step $1$ to step $i-1$ as $\bar{\mathbf{z}}_{[1,i-1]}$.
The final query $\mathbf{q}_i$ is constructed by fusing the global intent, the current textual rationale, and, when available, the previous latent state as:
\begin{equation}
\mathbf{q}_i = \mathrm{Average}\left( \mathbf{u}, \mathbf{q}^{\text{text}}_{[1,i]}, \mathbb{I}[i \textgreater 1] \cdot \bar{\mathbf{z}}_{[1,i-1]} \right).
\end{equation}
Finally, the teacher computes cosine similarities between $\mathbf{q}_i$ and each candidate feature in the refined pool $\mathbf{C}'_i$, and selects the top-$K$ most relevant features to form the supervision set $\mathbf{Z}_i$.

\subsection{Two-stage Learning}
\label{training}
We train the model using a two-stage pipeline that progressively instills interleaved latent reasoning capabilities using constructed supervision.

\paragraph{Stage 1: Interleaved Text-Latent Joint Supervision}
In the first stage, we enforce precise perceptual modeling. The teacher-selected features $\mathbf{Z}_i$ are used as teacher-forced inputs and supervision for the $K$ \texttt{<|latent\_pad|>} tokens at reasoning step $i$. The model is optimized with a joint loss: a standard cross-entropy loss $\mathcal{L}_{\text{CE}}$ for text tokens, and a latent alignment loss that forces the student's hidden state $\mathbf{h}_{t-1}$ to match the teacher's selected feature $\mathbf{z}_{t}$.
\begin{equation}
\begin{aligned}
    \mathcal{L}_{\text{S1}} & = \mathcal{L}_{\text{CE}}(\mathcal{X}_{\text{text}}) +  \\ &\lambda_{\text{sim}} \cdot \frac{1}{\sum_i K} \sum_i \sum_{t \in \mathcal{T}_i} \Big(1 - \cos\big(\mathbf{h}_{t-1},\, \mathbf{z}_{t}\big)\Big),
\end{aligned}
\end{equation}
where $\mathcal{T}_i$ is the indices of the latent tokens at reasoning step $i$, $\mathcal{X}_{\text{text}}$ represents all textual tokens, and $\lambda_{\text{sim}}$ balances the two objectives.

\paragraph{Stage 2: Text-Only Supervision with Latent Relaxation}
In the second stage, we relax the strict alignment constraint to allow the model to freely explore the latent reasoning process and use latent states as internal priors for subsequent tokens. 
We remove the latent alignment loss and feed self-generated hidden state as the input for the next latent position, optimizing only the textual part.
\begin{equation}
    \mathcal{L}_{\text{S2}} = \mathcal{L}_{\text{CE}}(\mathcal{X}_{\text{text}}),
\end{equation}

\begin{table*}[t]
\centering
\setlength{\tabcolsep}{3pt}
\resizebox{0.92\textwidth}{!}{
\begin{tabular}{ll ccccc c ccccc c}
\toprule
\multirow{2}{*}{\textbf{Methods}} 
& \multirow{2}{*}{\textbf{Paradigm}} 
& \multicolumn{5}{c}{\textbf{COMT}} & \multirow{2}{*}{\textbf{VSP}}
& \multicolumn{5}{c}{\textbf{COMT}} & \multirow{2}{*}{\textbf{VSP}} \\
\cmidrule(lr){3-7} \cmidrule(lr){9-13}

& 
& Creation & Deletion & Selection & Update & \textbf{Avg.} & 
& Creation & Deletion & Selection & Update & \textbf{Avg.} & \\
\midrule
\textbf{Backbones} & & \multicolumn{6}{c}{\textit{Qwen2.5-VL-7B}} & \multicolumn{6}{c}{\textit{Qwen3-VL-8B}} \\
\midrule

\rowcolor[HTML]{E1E1E1}
\multicolumn{2}{l}{\textit{Standard Baselines}} 
& \multicolumn{6}{c}{} 
& \multicolumn{6}{c}{} \\

Zero-shot 
& Direct Ans.
& 68.0 & 38.0 & 35.0 & 14.0 & 38.8 & 6.0
& \textbf{89.0} & 28.0 & 10.0 & 21.0 & 37.0 & 19.0 \\

Direct-FT 
& Direct Ans.
& 52.0 & 60.0 & 51.0 & 49.0 & 53.0 & 72.0
& 89.0 & 67.0 & 49.0 & 53.0 & 64.5 & 60.8 \\

CoT-FT 
& Text CoT
& \textbf{80.0} & 52.0 & 45.0 & 46.0 & 55.8 & 47.0
& 83.0 & 62.0 & 49.0 & 44.0 & 59.8 & 61.8 \\

\midrule

\multicolumn{14}{l}{\cellcolor[HTML]{E1E1E1}\textit{Latent Reasoning}} \\

\multicolumn{14}{l}{\cellcolor[gray]{0.97}\quad \textit{Stage 1: Latent Alignment}} \\

Mirage 
& Single-step
& 53.0 & 54.0 & 45.0 & 42.0 & 48.5 & 65.8
& 81.0 & 58.0 & 43.0 & 50.0 & 58.0 & 71.3 \\

\cellcolor{ourblue}ILVR (Ours)
& \cellcolor{ourblue}Interleaved
& \cellcolor{ourblue}69.0 & \cellcolor{ourblue}66.0 & \cellcolor{ourblue}46.0 & \cellcolor{ourblue}47.0 & \cellcolor{ourblue}57.0 & \cellcolor{ourblue}77.3
& \cellcolor{ourblue}{84.0} & \cellcolor{ourblue}{63.0} & \cellcolor{ourblue}{57.0} & \cellcolor{ourblue}{55.0} & \cellcolor{ourblue}{64.8} & \cellcolor{ourblue}{75.0} \\

\multicolumn{14}{l}{\cellcolor[gray]{0.97}\quad \textit{Stage 2: Latent Relaxation}} \\

Mirage 
& Single-step
& 65.0 & 62.0 & 47.0 & 50.0 & 56.0 & 76.0
& 84.0 & 66.0 & 54.0 & 57.0 & 65.3 & 78.3 \\

\cellcolor{ourblue}\textbf{ILVR (Ours)}
& \cellcolor{ourblue}\textbf{Interleaved}
& \cellcolor{ourblue}71.0 & \cellcolor{ourblue}\textbf{68.0} & \cellcolor{ourblue}\textbf{53.0} & \cellcolor{ourblue}\textbf{51.0} & \cellcolor{ourblue}\textbf{60.8} & \cellcolor{ourblue}\textbf{81.5}
& \cellcolor{ourblue}87.0 & \cellcolor{ourblue}\textbf{73.0} & \cellcolor{ourblue}\textbf{60.0} & \cellcolor{ourblue}\textbf{62.0} & \cellcolor{ourblue}\textbf{70.5} & \cellcolor{ourblue}\textbf{82.8} \\

\bottomrule
\end{tabular}}
\caption{\textbf{IID performance comparison on COMT and VSP.} Creation, Deletion, Selection, and Update denote COMT subtasks. Backbone differences are explicitly indicated in the \textit{Standard Baselines} header row, while latent reasoning methods are evaluated under the same column layout. ``Direct Ans.'' and ``Text CoT'' denote direct answer generation and text-only CoT, respectively. \textbf{Bold} indicates the best result. Accuracy (\%) is reported.}
\label{tab:main_results_iid_combined_v3}
\end{table*}

\section{Experiments}
\label{sec:experiments}
\subsection{Experimental Setup}

\paragraph{Datasets} We evaluate ILVR under both in-distribution (IID) and out-of-distribution (OOD) settings. IID evaluation follows the standard splits of COMT~\citep{Cheng2024CoMTAN} and VSP~\citep{Wu2024VSPAT}. For OOD evaluation, models are trained on a 10k subset of Zebra-CoT~\citep{Li2025ZebraCoTAD} spanning scientific, visual logic, and 3D reasoning tasks, then evaluated on EMMA BENCH~\citep{Hao2025CanMR}, VisuLogic~\citep{Xu2025VisuLogicAB}, and held-out Zebra-CoT 2D visual reasoning tasks. The OOD setting is characterized by task-type mismatch: Zebra-CoT science focuses on physics and graph problems, whereas EMMA BENCH additionally covers mathematics, chemistry, and coding; Zebra-CoT visual logic centers on maze- and game-like tasks, while VisuLogic targets positional, quantitative, and stylistic reasoning. Controlled comparisons are conducted on both \texttt{Qwen2.5-VL-7B} and \texttt{Qwen3-VL-8B}~\citep{Bai2025Qwen25VLTR} backbones to demonstrate generalization.

\paragraph{Baselines} We compare ILVR against three categories of baselines: (1) Standard baselines, including Zero-shot, direct answer fine-tuning  (Direct-FT) and CoT fine-tuning (CoT-FT). (2) Single-step latent reasoning methods, i.e., Mirage~\citep{Yang2025MachineMI} and LVR~\citep{Li2025LatentVR}. We report Mirage as the representative baseline in main tables, as LVR operates on pre-defined bounding boxes and models only static visual states, making it incompatible with dynamically evolving reasoning scenarios in our benchmarks.

We report additional experiments against LVR on bounding-box–annotated data in Appendix~\ref{app:additional_results}.
(3) SOTA reasoning models (OOD only) with extensive reinforcement learning (RL), including VisionR1~\citep{Huang2025VisionR1IR} and PixelReasoner~\citep{Su2025PixelRI}. We also compare against Bagel-Zebra~\citep{Li2025ZebraCoTAD}, a Bagel~\citep{Deng2025EmergingPI} variant fine-tuned on the complete 180k Zebra-CoT dataset to enhance reasoning capabilities. We use official checkpoints for specialized models and fine-tune all others on the same datasets as ILVR with identical implementation settings for both backbones. Notably, we omit Bagel-Zebra from Zebra-CoT OOD evaluation because it was trained on the full dataset, making the test set in-distribution and unsuitable for OOD comparison.

\begin{table*}[th]
\centering
\definecolor{ourblue}{RGB}{235, 242, 250}
\setlength{\tabcolsep}{4pt} 
\resizebox{\textwidth}{!}{
\begin{tabular}{ll ccccc c cccc c ccc c c}
\toprule
\multirow{2}{*}{\textbf{Model}} & \multirow{2}{*}{\textbf{Paradigm}} & \multicolumn{5}{c}{\textbf{EMMA BENCH}} & & \multicolumn{4}{c}{\textbf{VisuLogic}} & & \multicolumn{3}{c}{\textbf{Zebra-CoT} (OOD)} & & \multirow{2}{*}{\textbf{Total}} \\
\cmidrule(lr){3-7} \cmidrule(lr){9-12} \cmidrule(lr){14-16} 
 &  & Chem. & Code & Math & Phys. & \textbf{Avg.} & & Pos. & Quant. & Style & \textbf{Avg.} & & Jigsaw & Search & \textbf{Avg.} & &  \\ 
\midrule

\rowcolor[HTML]{E1E1E1}\multicolumn{18}{l}{\textit{SOTA Reasoning Models (Official Checkpoints - No Task-specific Fine-tuning)}} \\
VisionR1 & Reasoning & 15.0 & 30.0 & 32.0 & 20.0 & 24.3 & & 18.0 & 13.0 & 14.0 & 15.2 & & \textbf{25.0} & 65.0 & 45.0 & & 29.5 \\
PixelReasoner & Tool-use & 19.0 & 22.0 & 26.0 & 27.0 & 23.5 & & 18.0 & 16.0 & 29.0 & 23.4 & & 18.0 & 73.0 & 45.5 & & 31.6 \\
Bagel-Zebra & Unified & 23.0 & 28.0 & 29.0 & 32.0 & 28.0 & & 28.0 & \textbf{39.0} & 21.0 & 28.9 & & - & - & - & & - \\
\midrule

\rowcolor[HTML]{E1E1E1}\multicolumn{18}{l}{\textit{Standard Baselines (Fine-tuned on Zebra-CoT 10k subset)}} \\
Zero-shot & Direct Ans. & 18.0 & 25.0 & 28.0 & 33.0 & 26.0 & & \textbf{29.0} & 24.0 & 27.0 & 26.6 & & 23.0 & 65.0 & 44.0 & & 32.8 \\
Direct-FT & Direct Ans. & 16.0 & 27.0 & 28.0 & 32.0 & 25.8 & & 25.0 & 23.0 & 23.0 & 23.8 & & 17.0 & 73.0 & 45.0 & & 32.3 \\
CoT-FT & Text CoT & 21.0 & 26.0 & 33.0 & 31.0 & 27.8 & & 27.0 & 23.0 & 28.0 & 25.9 & & 21.5 & 68.5 & 45.0 & & 33.6 \\ 

\cmidrule(lr){1-18}
\rowcolor[HTML]{E1E1E1}\multicolumn{18}{l}{\textit{Latent Reasoning (Fine-tuned on Zebra-CoT 10k subset)}} \\
\rowcolor[gray]{0.97} \multicolumn{18}{l}{\quad \textit{Stage 1: Latent Alignment}} \\
Mirage & Single-step & 13.0 & 21.0 & 30.0 & \textbf{37.0} & 25.3 & & 25.0 & 24.0 & 21.0 & 23.4 & & 16.0 & 71.0 & 43.5 & & 31.5 \\
\rowcolor{ourblue} ILVR (Ours) & Interleaved & 23.0 & 26.0 & 34.0 & 35.0 & 29.5 & & 26.0 & 23.0 & 24.0 & 24.5 & & 20.5 & 74.5 & 47.5 & & 34.8 \\ 

\rowcolor[gray]{0.97} \multicolumn{18}{l}{\quad \textit{Stage 2: Latent Relaxation}} \\
Mirage & Single-step & 15.0 & 25.0 & \textbf{35.0} & 33.0 & 27.0 & & 24.0 & 26.0 & 30.0 & 26.6 & & 20.0 & \textbf{74.5} & 47.3 & & 34.3 \\
\rowcolor{ourblue} \textbf{ILVR (Ours)} & \textbf{Interleaved} & \textbf{31.0} & \textbf{35.0} & 34.0 & 33.0 & \textbf{33.3} & & 27.0 & 30.0 & \textbf{31.0} & \textbf{29.3} & & 22.5 & 73.0 & \textbf{47.8} & & \textbf{37.5} \\ 

\bottomrule
\end{tabular}
}
\caption{\textbf{Generalization evaluation on three OOD benchmarks: EMMA BENCH, VisuLogic, and Zebra-CoT.} The table compares state-of-the-art RL-based reasoning models using official checkpoints, standard baselines fine-tuned on Zebra-CoT (10k subset), and our ILVR. 
\textbf{Bold} indicates the best result within each column.
As Bagel-Zebra is trained on the full Zebra-CoT dataset (180k), making the Zebra-CoT test set in-distribution for this model. We thus omit its score in the OOD Zebra-CoT column to ensure a fair comparison. Accuracy (\%) is reported.}
\label{tab:combined_results_final}
\end{table*}

\paragraph{Implementation Details.}
We optimize all models using AdamW with a learning rate of 1e-5, a cosine learning-rate scheduler, and a fixed random seed of 42. For IID tasks (COMT and VSP), training is conducted for 15 epochs. In the OOD setting, models are fine-tuned on the 10k Zebra-CoT subset for 2 epochs with a target group size $L=784$ for adaptive feature grouping. Across all experiments, we set the latent token size to $K=8$, the alignment weight $\lambda_{\text{sim}}=1$, and the EMA decay to $\tau=0.999$. \texttt{Qwen2.5-VL-72B} serves as the judge model for open-ended evaluations.

\subsection{Main Results}
\label{sec:results}

Table~\ref{tab:main_results_iid_combined_v3} reports in-distribution results on COMT and VSP. ILVR consistently outperforms standard baselines, including Zero-shot, Direct-FT, CoT-FT, and the single-step latent method Mirage across both backbones. 
With the \texttt{Qwen2.5-VL-7B} backbone, ILVR achieves 60.8\% accuracy on COMT and 81.5\% on VSP, surpassing Mirage by 4.8\% and 5.5\%, respectively. 
When scaling to the stronger \texttt{Qwen3-VL-8B}, ILVR maintains this significant advantage, reaching 70.5\% on COMT (+5.2\%) and 82.8\% on VSP (+4.5\%). These results confirm that interleaved text–latent reasoning yields consistent and backbone-agnostic benefits, leading to stronger overall performance.

Table~\ref{tab:combined_results_final} shows that these gains transfer to OOD evaluation where ILVR consistently outperforms standard baseline and the latent method Mirage across all benchmarks, achieving an average improvement of 3.2\% over Mirage. ILVR also surpasses recent state-of-the-art multimodal reasoning models VisionR1 and PixelReasoner despite their use of more stochastic reinforcement learning. In terms of average accuracy, ILVR exceeds VisionR1 by 8.0\% and PixelReasoner by 5.9\%. We further compare ILVR with Bagel-Zebra trained on the full Zebra-CoT dataset with 180k samples. ILVR is trained on only a 10k subset yet still outperforms Bagel-Zebra on EMMA BENCH and VisuLogic. Results on Zebra-CoT OOD are omitted for Bagel-Zebra, as its test split becomes in-distribution.
\begin{table}[h!]
\centering
\resizebox{\columnwidth}{!}{
\setlength{\tabcolsep}{3pt}
\begin{tabular}{l c c cccc}
\toprule
\multirow{2}{*}{\makecell{\textbf{Reasoning}\\\textbf{Paradigm}}}
& 
\multirow{2}{*}{\makecell{\textbf{Perception}\\\textbf{Mechanism}}} &
& \multicolumn{4}{c}{\textbf{Accuracy}} \\
\cmidrule(lr){4-7}
& & & VisLog & EMMA & Zebra & \textbf{Total} \\
\midrule
Single-step & Mean Pooling & (Mirage) & 23.4 & 25.3 & 43.5 & 31.5 \\
Single-step & Selective & / & 24.1 & 26.3 & 44.5 & 32.4 \\
\rowcolor{ourblue} Interleaved & Selective & (\textbf{ILVR})
& \textbf{24.5} & \textbf{29.5} & \textbf{47.5} & \textbf{34.8} \\
\bottomrule
\end{tabular}
}
\caption{\textbf{Ablation of interleaved paradigm and selection perception mechanism} against mean pooling and single-step setup (Mirage). Accuracy (\%) is reported.}
\label{tab:ablation_components}
\end{table}
\begin{figure}[h!]
    \centering
    \includegraphics[width=0.8\linewidth]{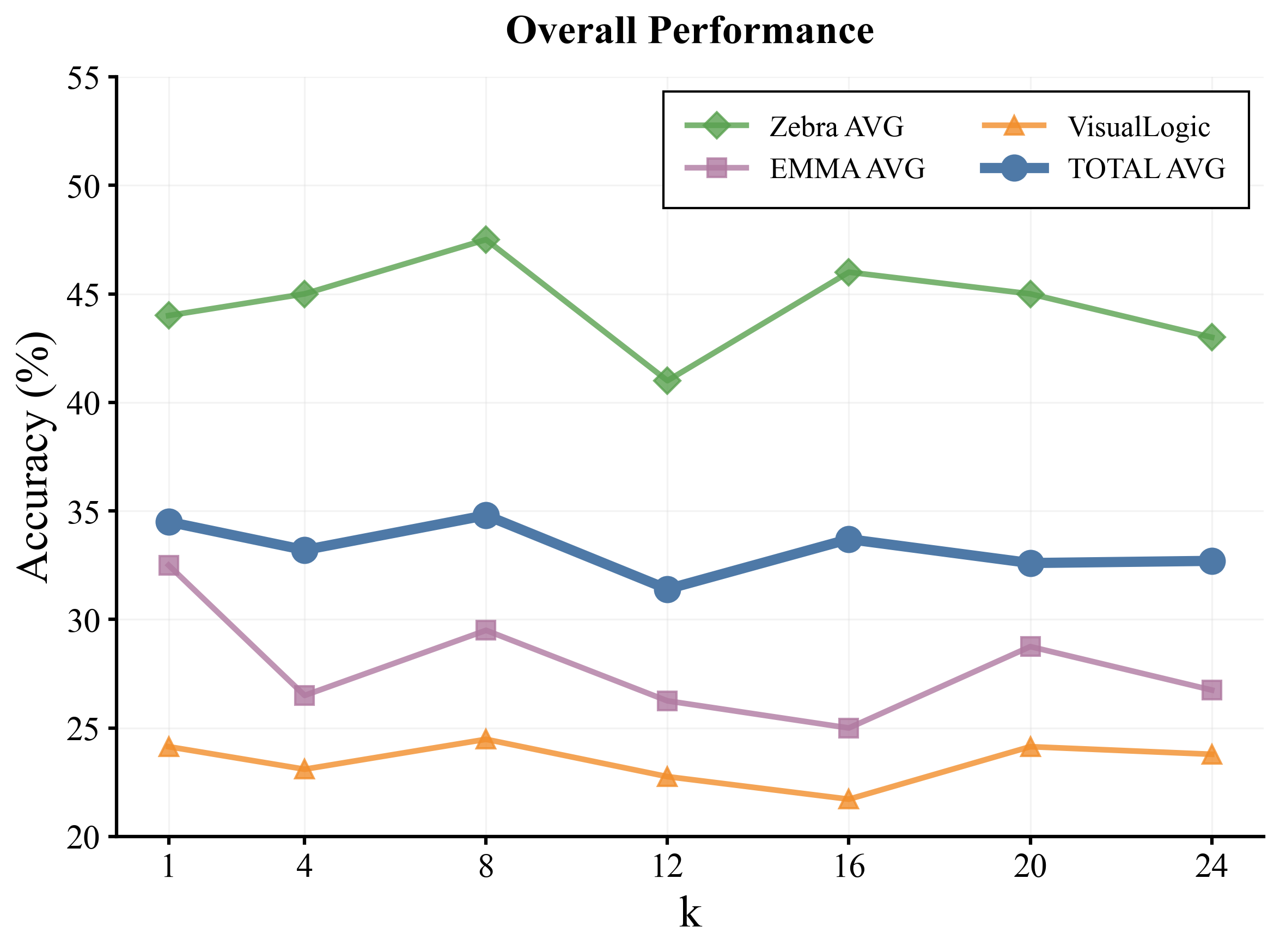}
    \caption{\textbf{Impact of latent size $K$.} Performance trends across VisuLogic, EMMA, and Zebra-CoT, as well as the overall average, as the number of latent tokens $K$ varies. $\lambda_{\text{sim}}$ is fixed at 1.0. $K=8$ yields the most robust performance across benchmarks.}
    \label{fig:ablation_k}
\end{figure}

\begin{table}[h!]
\centering

\small
\setlength{\tabcolsep}{5pt}
\resizebox{0.7\columnwidth}{!}{
\begin{tabular}{l cccc}
\toprule
\multirow{2}{*}{$\boldsymbol{\lambda}_{\textbf{sim}}$} & \multicolumn{4}{c}{\textbf{Accuracy}} \\
\cmidrule(lr){2-5}
 & VisLog & EMMA & Zebra & \textbf{Total} \\
\midrule
0.1 & 23.4 & 25.8 & 44.0 & 31.8 \\
0.5 & 20.0 & \textbf{30.5} & 45.8 & 33.3 \\
\rowcolor{ourblue} \textbf{1} \textbf{(ILVR)} & \textbf{24.5} & 29.5 & 47.5 & \textbf{34.8} \\
2 & 21.7 & 27.8 & 42.5 & 31.6 \\
10 & 21.4 & 27.5 & \textbf{48.5} & 33.6 \\
\bottomrule
\end{tabular}}
\caption{\textbf{Sensitivity to alignment loss weight $\lambda_{\text{sim}}$.} We report the average accuracy (\%) across benchmarks. $\lambda_{\text{sim}}$ represents the relative weight of the alignment loss relative to the text generation loss. }
\label{tab:ablation_lambda}
\end{table}

\begin{figure*}[t!]
    \centering
    \includegraphics[width=0.9\linewidth]{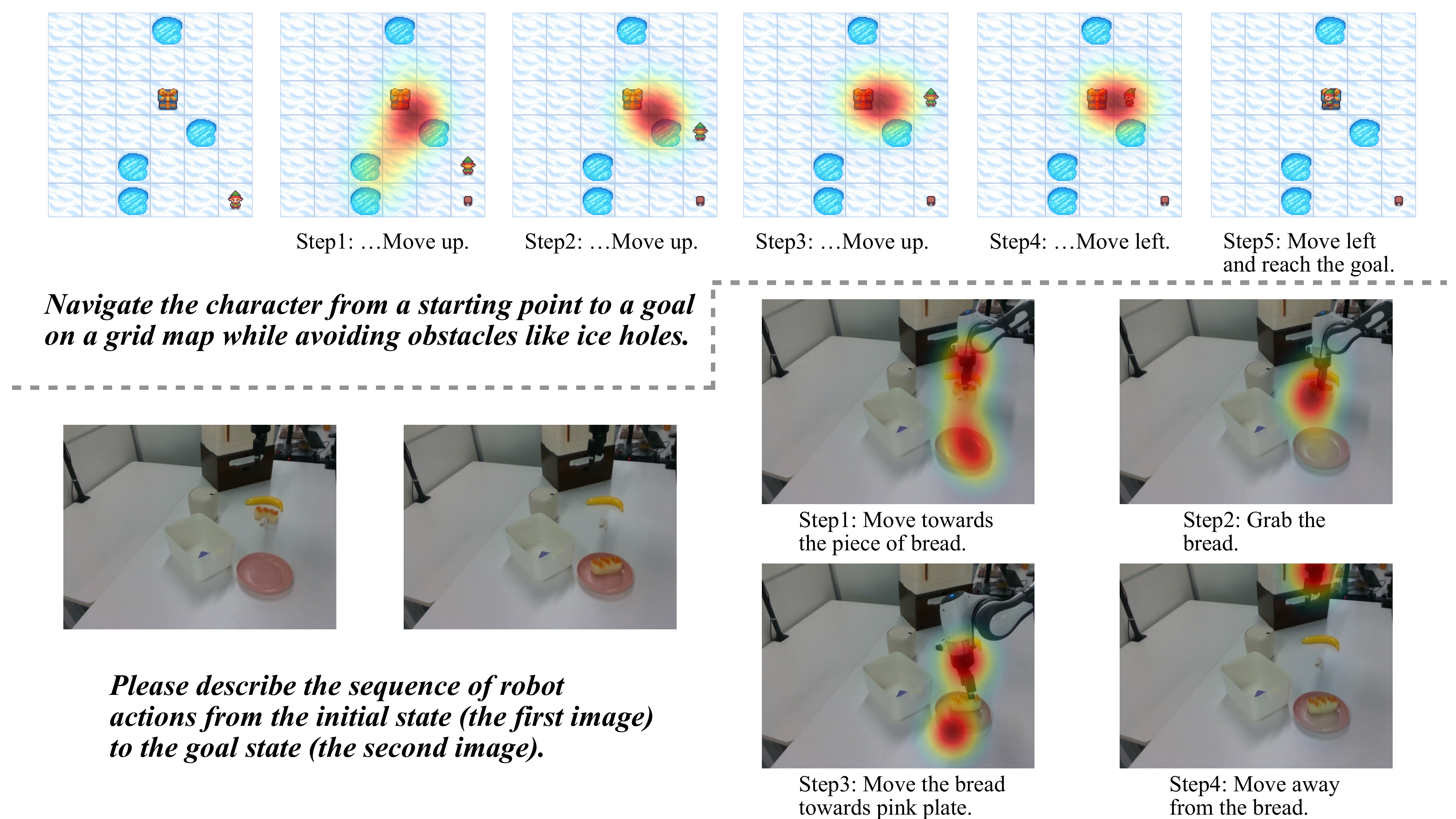}
    \caption{
        \textbf{Visualization of dynamic latent modeling.} 
        Heatmaps depict the Gaussian-smoothed aggregation of relevant image patches for $K=8$ generated latents.
        \textbf{Top} (Navigation): Latents sequentially track the character's planned path.
        \textbf{Bottom} (Robotic Manipulation): Visual attention shifts from the object (bread) to the target (plate) during the task.
        These confirm precise alignment between generated latents and the step-wise reasoning context.
    }
    \label{fig:visualization}
\end{figure*}

\begin{figure}[h!]
    \centering
    \includegraphics[width=0.95\linewidth]{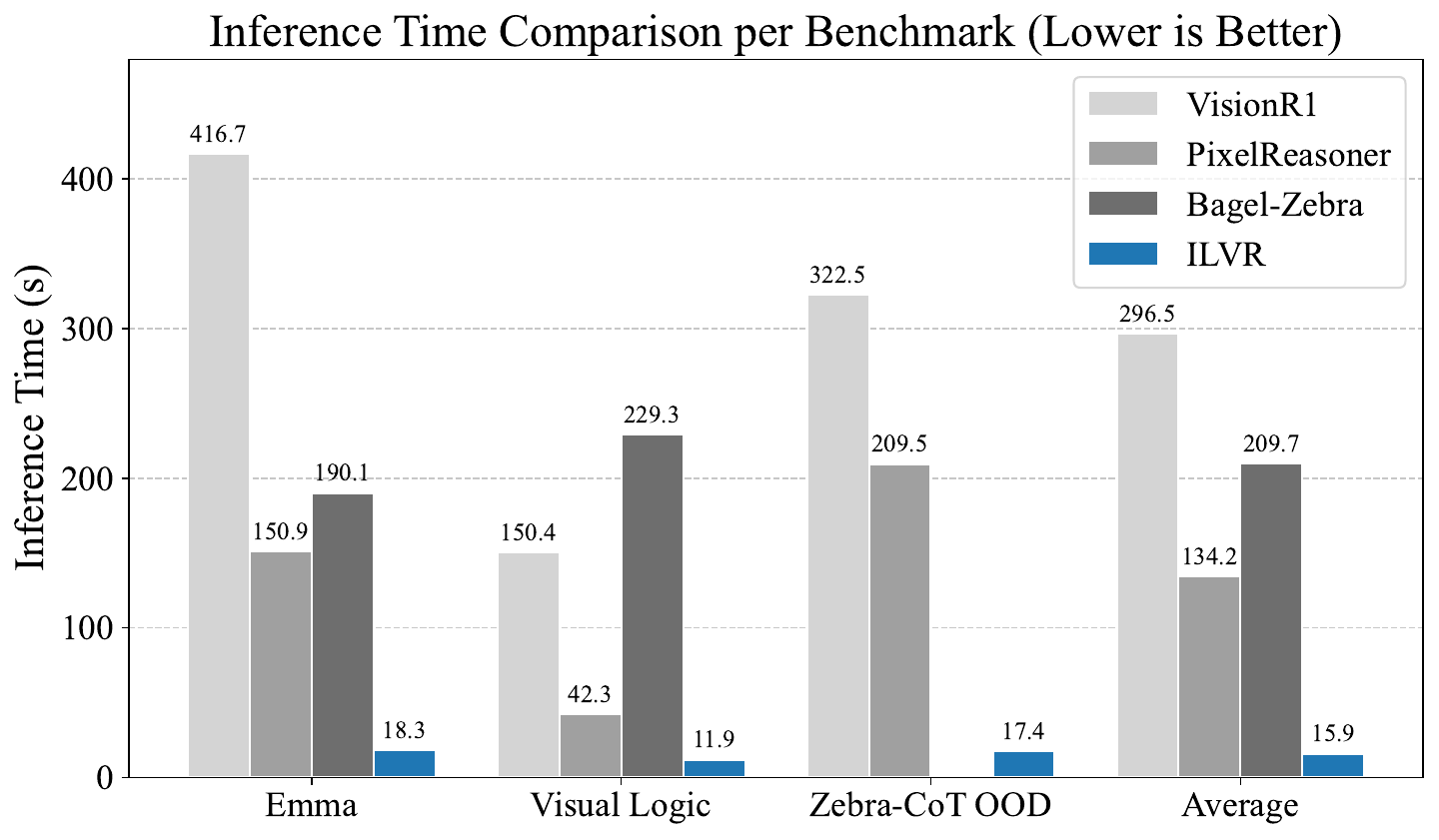}
    \caption{\textbf{Comparison of average inference time per sample.} We report the latency averaged across EMMA, VisuLogic, and Zebra-CoT benchmarks.}
    \label{fig:efficiency}
\end{figure}

\subsection{Ablation Study}
\label{sec:ablation}
We conduct ablations based on Stage 1 training and OOD benchmarks to investigate the contribution of interleaved reasoning and selective perception in our ILVR, and analyze the impact of different latent sizes $K$ and the alignment weights $\lambda_{\text{sim}}$.

\paragraph{Interleaved \& Selective Design.}
Table~\ref{tab:ablation_components} shows that replacing mean pooling with teacher-guided selective perceptual modeling improves the overall accuracy from 31.5\% to 32.4\%. Adding the interleaved reasoning paradigm yields further gains to 34.8\%. These results suggest that selective perception improves the quality of latent supervision, and interleaved latent updates boost performance by explicitly modeling evolving reasoning states.

\paragraph{Latent size $K$.}
Fig.~\ref{fig:ablation_k} reports performance under different latent sizes $K$, where $K=8$ yields the best overall results. This indicates that a moderate latent budget is sufficient to capture step-specific perceptual evidence, while smaller $K$ limits representational capacity and larger $K$ introduces redundant latent content that weakens step-wise updates. We therefore use $K=8$ in all experiments.

\paragraph{Alignment weight $\lambda_{\text{sim}}$.}
Table~\ref{tab:ablation_lambda} reports sensitivity to $\lambda_{\text{sim}}$ and shows that $\lambda_{\text{sim}}=1$ yields the best accuracy. Smaller values weaken latent supervision and perceptual grounding, while larger values over-constrain latent representations and hinder adaptation to subsequent reasoning steps. This supports $\lambda_{\text{sim}}=1$ as an effective trade-off between perceptual alignment and reasoning flexibility.

\subsection{Analysis}
\label{sec:analysis}
\paragraph{Efficiency.}
Fig.~\ref{fig:efficiency} reports the average inference time per sample on a same NVIDIA H200 GPU, averaged across EMMA BENCH, VisuLogic, and Zebra-CoT OOD. ILVR achieves substantially lower latency than competing methods, running orders of magnitude faster ($\times8 \sim \times18$ speedup) than VisionR1, PixelReasoner, and Bagel-Zebra.

The key reason is that ILVR performs multi-step reasoning by updating compact latent states, which bypasses repeated pixel-level processing and intermediate image generation that dominate the runtime of these baselines. These results confirm that ILVR provides an efficient alternative to costly long-context or tool-based reasoning methods.

\paragraph{Qualitative visualization.}
Fig.~\ref{fig:visualization} visualizes Gaussian-smoothed aggregation of relevant patches derived from attention weights for $K=8$ generated latents. In the navigation example, attention evolves step by step with the planned actions. Early steps focus on both the goal and nearby ice holes to ensure safe planning, while later steps concentrate almost entirely on the goal once the path is clear. In the robotic manipulation example, attention concentrates on the bread during approaching and grasping, then shifts toward the plate during subgoal placement, and finally moves away after completion. This step-aligned evolution suggests that the latents are conditioned on the evolving reasoning context and provide subgoal-adaptive localized visual cues, which helps subsequent text generation remain grounded in the correct regions.

\section{Conclusion}

In this paper, we introduce Interleaved Latent Visual Reasoning (ILVR) to unify dynamic state evolution with precise perceptual modeling. Unlike single-step methods that bypass intermediate verification, ILVR interleaves textual generation with evolving latent representations to track reasoning states without costly pixel-level re-encoding. Our momentum teacher-guided selection mechanism distills step-specific visual cues, avoiding feature over-compression. Experiments confirm that ILVR significantly outperforms single-step latent methods, validating dynamic latent reasoning as a scalable path for multimodal intelligence.

\section*{Limitations}
Despite ILVR's robust performance, three limitations remain for future work. 
First, while theoretically model-agnostic, our experiments currently focus on Qwen-VL backbones; validating the framework across diverse architectures and larger parameter scales is a necessary next step. 
Second, integrating Reinforcement Learning (RL) to directly optimize latent trajectories could further enhance multi-step planning capabilities. 
Finally, although attention maps provide insight, the generated latent representations are not directly human-readable; exploring decoding mechanisms to project these states back into pixel space remains an open challenge for better interpretability.

\section*{Use of AI Assistants}
In adherence to the ACL Publication Ethics Policy, we did not employ AI assistants to generate the initial draft of this paper. We used AI assistants such as GPT-5.2 and Gemini3-Pro exclusively at the sentence level to enhance our writing quality and correct grammatical errors.

\bibliographystyle{acl_natbib}
\bibliography{main}

\begin{thebibliography}{30}
\providecommand{\natexlab}[1]{#1}

\bibitem[{Bai et~al.(2025{\natexlab{a}})Bai, Chen, Liu, Wang, Ge, Song, Dang, Wang, Wang, Tang, Zhong, Zhu, Yang, Li, Wan, Wang, Ding, Fu, Xu, Ye, Zhang, Xie, Cheng, Zhang, Yang, Xu, and Lin}]{Bai2025Qwen25VLTR}
Shuai Bai, Keqin Chen, Xuejing Liu, Jialin Wang, Wenbin Ge, Sibo Song, Kai Dang, Peng Wang, Shijie Wang, Jun Tang, Humen Zhong, Yuanzhi Zhu, Mingkun Yang, Zhaohai Li, Jianqiang Wan, Pengfei Wang, Wei Ding, Zheren Fu, Yiheng Xu, and 8 others. 2025{\natexlab{a}}.
\newblock \href {https://api.semanticscholar.org/CorpusID:276449796} {Qwen2.5-vl technical report}.
\newblock \emph{ArXiv}, abs/2502.13923.

\bibitem[{Bai et~al.(2025{\natexlab{b}})Bai, Li, Liu, Tang, Zhang, Sun, Chu, and Tang}]{Bai2025UniVGR1RG}
Sule Bai, Mingxing Li, Yong Liu, Jing Tang, Haoji Zhang, Lei Sun, Xiangxiang Chu, and Yansong Tang. 2025{\natexlab{b}}.
\newblock \href {https://api.semanticscholar.org/CorpusID:278769702} {Univg-r1: Reasoning guided universal visual grounding with reinforcement learning}.
\newblock \emph{ArXiv}, abs/2505.14231.

\bibitem[{Cheng and Durme(2024)}]{Cheng2024CompressedCO}
Jeffrey Cheng and Benjamin~Van Durme. 2024.
\newblock \href {https://api.semanticscholar.org/CorpusID:274789675} {Compressed chain of thought: Efficient reasoning through dense representations}.
\newblock \emph{ArXiv}, abs/2412.13171.

\bibitem[{Cheng et~al.(2024)Cheng, Chen, Zhang, Fei, Feng, Che, Li, and Qin}]{Cheng2024CoMTAN}
Zihui Cheng, Qiguang Chen, Jin Zhang, Hao Fei, Xiaocheng Feng, Wanxiang Che, Min Li, and Libo Qin. 2024.
\newblock \href {https://api.semanticscholar.org/CorpusID:274789454} {Comt: A novel benchmark for chain of multi-modal thought on large vision-language models}.
\newblock \emph{ArXiv}, abs/2412.12932.

\bibitem[{Chern et~al.(2024)Chern, Su, Ma, and Liu}]{Chern2024ANOLEAO}
Ethan Chern, Jiadi Su, Yan Ma, and Pengfei Liu. 2024.
\newblock \href {https://api.semanticscholar.org/CorpusID:271050462} {Anole: An open, autoregressive, native large multimodal models for interleaved image-text generation}.
\newblock \emph{ArXiv}, abs/2407.06135.

\bibitem[{Deng et~al.(2025)Deng, Zhu, Li, Gou, Li, Wang, Zhong, Yu, Nie, Song, Guang, and Fan}]{Deng2025EmergingPI}
Chaorui Deng, Deyao Zhu, Kunchang Li, Chenhui Gou, Feng Li, Zeyu Wang, Shu Zhong, Weihao Yu, Xiaonan Nie, Ziang Song, Shi Guang, and Haoqi Fan. 2025.
\newblock \href {https://api.semanticscholar.org/CorpusID:278768720} {Emerging properties in unified multimodal pretraining}.
\newblock \emph{ArXiv}, abs/2505.14683.

\bibitem[{Fu et~al.(2025)Fu, Liu, Yang, Corring, Lu, Yang, Roth, Flor{\^e}ncio, and Zhang}]{Fu2025ReFocusVE}
Xingyu Fu, Minqian Liu, Zhengyuan Yang, John Corring, Yijuan Lu, Jianwei Yang, Dan Roth, Dinei A.~F. Flor{\^e}ncio, and Cha Zhang. 2025.
\newblock \href {https://api.semanticscholar.org/CorpusID:275405594} {Refocus: Visual editing as a chain of thought for structured image understanding}.
\newblock \emph{ArXiv}, abs/2501.05452.

\bibitem[{Hao et~al.(2024)Hao, Sukhbaatar, Su, Li, Hu, Weston, and Tian}]{Hao2024TrainingLL}
Shibo Hao, Sainbayar Sukhbaatar, DiJia Su, Xian Li, Zhiting Hu, Jason~E. Weston, and Yuandong Tian. 2024.
\newblock \href {https://api.semanticscholar.org/CorpusID:274610816} {Training large language models to reason in a continuous latent space}.
\newblock \emph{ArXiv}, abs/2412.06769.

\bibitem[{Hao et~al.(2025)Hao, Gu, Wang, Li, Yang, Wang, and Cheng}]{Hao2025CanMR}
Yunzhuo Hao, Jiawei Gu, Huichen~Will Wang, Linjie Li, Zhengyuan Yang, Lijuan Wang, and Yu~Cheng. 2025.
\newblock \href {https://api.semanticscholar.org/CorpusID:275405458} {Can mllms reason in multimodality? emma: An enhanced multimodal reasoning benchmark}.
\newblock \emph{ArXiv}, abs/2501.05444.

\bibitem[{He et~al.(2019)He, Fan, Wu, Xie, and Girshick}]{He2019MomentumCF}
Kaiming He, Haoqi Fan, Yuxin Wu, Saining Xie, and Ross~B. Girshick. 2019.
\newblock \href {https://api.semanticscholar.org/CorpusID:207930212} {Momentum contrast for unsupervised visual representation learning}.
\newblock \emph{2020 IEEE/CVF Conference on Computer Vision and Pattern Recognition (CVPR)}, pages 9726--9735.

\bibitem[{Hu et~al.(2024)Hu, Shi, Fu, Roth, Ostendorf, Zettlemoyer, Smith, and Krishna}]{Hu2024VisualSS}
Yushi Hu, Weijia Shi, Xingyu Fu, Dan Roth, Mari Ostendorf, Luke~S. Zettlemoyer, Noah~A. Smith, and Ranjay Krishna. 2024.
\newblock \href {https://api.semanticscholar.org/CorpusID:270440440} {Visual sketchpad: Sketching as a visual chain of thought for multimodal language models}.
\newblock \emph{ArXiv}, abs/2406.09403.

\bibitem[{Huang et~al.(2025{\natexlab{a}})Huang, Jia, Zhai, Cao, Ye, Zhao, Xu, Hu, and Lin}]{Huang2025VisionR1IR}
Wenxuan Huang, Bohan Jia, Zijie Zhai, Shaoshen Cao, Zheyu Ye, Fei Zhao, Zhe Xu, Yao Hu, and Shaohui Lin. 2025{\natexlab{a}}.
\newblock \href {https://api.semanticscholar.org/CorpusID:276902576} {Vision-r1: Incentivizing reasoning capability in multimodal large language models}.
\newblock \emph{ArXiv}, abs/2503.06749.

\bibitem[{Huang et~al.(2025{\natexlab{b}})Huang, Ji, Rajan, Cai, Xiao, Hu, and Lee}]{Huang2025VisualToolAgentA}
Zeyi Huang, Yuyang Ji, Anirudh~Sundara Rajan, Zefan Cai, Wen Xiao, Junjie Hu, and Yong~Jae Lee. 2025{\natexlab{b}}.
\newblock \href {https://api.semanticscholar.org/CorpusID:278910554} {Visualtoolagent (vista): A reinforcement learning framework for visual tool selection}.
\newblock \emph{ArXiv}, abs/2505.20289.

\bibitem[{Li et~al.(2025{\natexlab{a}})Li, Wang, Yue, Cai, Liu, Fu, Guo, Zhu, Sharan, Jia, Neiswanger, Huang, Goldstein, and Goldblum}]{Li2025ZebraCoTAD}
Ang Li, Charles~L. Wang, Kaiyu Yue, Zikui Cai, Ollie Liu, Deqing Fu, Peng Guo, Wang~Bill Zhu, Vatsal Sharan, Robin Jia, Willie Neiswanger, Furong Huang, Tom Goldstein, and Micah Goldblum. 2025{\natexlab{a}}.
\newblock \href {https://api.semanticscholar.org/CorpusID:280165703} {Zebra-cot: A dataset for interleaved vision language reasoning}.
\newblock \emph{ArXiv}, abs/2507.16746.

\bibitem[{Li et~al.(2025{\natexlab{b}})Li, Sun, Liu, Wang, Wu, Yu, Chen, Barsoum, Chen, and Liu}]{Li2025LatentVR}
Bangzheng Li, Ximeng Sun, Jiang Liu, Ze~Wang, Jialian Wu, Xiaodong Yu, Hao Chen, Emad Barsoum, Muhao Chen, and Zicheng Liu. 2025{\natexlab{b}}.
\newblock \href {https://api.semanticscholar.org/CorpusID:281675495} {Latent visual reasoning}.
\newblock \emph{ArXiv}, abs/2509.24251.

\bibitem[{Li et~al.(2024)Li, Zhang, Guo, Zhang, Li, Zhang, Zhang, Li, Liu, and Li}]{Li2024LLaVAOneVisionEV}
Bo~Li, Yuanhan Zhang, Dong Guo, Renrui Zhang, Feng Li, Hao Zhang, Kaichen Zhang, Yanwei Li, Ziwei Liu, and Chunyuan Li. 2024.
\newblock \href {https://api.semanticscholar.org/CorpusID:271719914} {Llava-onevision: Easy visual task transfer}.
\newblock \emph{ArXiv}, abs/2408.03326.

\bibitem[{Liu et~al.(2025)Liu, Wang, Ruan, Luo, Chen, Li, and Liu}]{Liu2025VisualAT}
Dairu Liu, Ziyue Wang, Minyuan Ruan, Fuwen Luo, Chi Chen, Peng Li, and Yang Liu. 2025.
\newblock \href {https://api.semanticscholar.org/CorpusID:278912198} {Visual abstract thinking empowers multimodal reasoning}.
\newblock \emph{ArXiv}, abs/2505.20164.

\bibitem[{Shao et~al.(2024{\natexlab{a}})Shao, Qian, Xiao, Song, Zong, Wang, Liu, and Li}]{Shao2024VisualCA}
Hao Shao, Shengju Qian, Han Xiao, Guanglu Song, Zhuofan Zong, Letian Wang, Yu~Liu, and Hongsheng Li. 2024{\natexlab{a}}.
\newblock \href {https://api.semanticscholar.org/CorpusID:271051212} {Visual cot: Advancing multi-modal language models with a comprehensive dataset and benchmark for chain-of-thought reasoning}.
\newblock \emph{Advances in Neural Information Processing Systems 37}.

\bibitem[{Shao et~al.(2024{\natexlab{b}})Shao, Qian, Xiao, Song, Zong, Wang, Liu, and Li}]{Shao2024VisualCU}
Hao Shao, Shengju Qian, Han Xiao, Guanglu Song, Zhuofan Zong, Letian Wang, Yu~Liu, and Hongsheng Li. 2024{\natexlab{b}}.
\newblock \href {https://api.semanticscholar.org/CorpusID:268681119} {Visual cot: Unleashing chain-of-thought reasoning in multi-modal language models}.
\newblock \emph{ArXiv}, abs/2403.16999.

\bibitem[{Shen et~al.(2025)Shen, Yan, Zhang, Hu, Du, and He}]{Shen2025CODICC}
Zhenyi Shen, Hanqi Yan, Linhai Zhang, Zhanghao Hu, Yali Du, and Yulan He. 2025.
\newblock \href {https://api.semanticscholar.org/CorpusID:276725056} {Codi: Compressing chain-of-thought into continuous space via self-distillation}.
\newblock \emph{ArXiv}, abs/2502.21074.

\bibitem[{Su et~al.(2025)Su, Wang, Ren, Lin, and Chen}]{Su2025PixelRI}
Alex Su, Haozhe Wang, Weiming Ren, Fangzhen Lin, and Wenhu Chen. 2025.
\newblock \href {https://api.semanticscholar.org/CorpusID:278789415} {Pixel reasoner: Incentivizing pixel-space reasoning with curiosity-driven reinforcement learning}.
\newblock \emph{ArXiv}, abs/2505.15966.

\bibitem[{Wang et~al.(2025{\natexlab{a}})Wang, Kang, Wang, Jiang, Li, Wu, Wang, Ran, Liang, Feng, and Xiao}]{Wang2025VGRVG}
Jiacong Wang, Zijiang Kang, Haochen Wang, Haiyong Jiang, Jiawen Li, Bohong Wu, Ya~Wang, Jiao Ran, Xiao Liang, Chao Feng, and Jun Xiao. 2025{\natexlab{a}}.
\newblock \href {https://api.semanticscholar.org/CorpusID:279391256} {Vgr: Visual grounded reasoning}.
\newblock \emph{ArXiv}, abs/2506.11991.

\bibitem[{Wang et~al.(2025{\natexlab{b}})Wang, Gao, Gu, Pu, Cui, Wei, Liu, Jing, Ye, Shao, Wang, Chen, Zhang, Yang, Wang, Wei, Yin, Li, Cui, Chen, Ding, Tian, Wu, Xie, Li, Yang, Duan, Wang, Hao, Li, Zhao, Duan, Deng, Fu, He, Wang, He, Shi, He, Xiong, Lv, Wu, Shao, Zhang, Deng, Qi, Ge, Guo, Zhang, Gu, Ouyang, Wang, Dou, Zhu, Lu, Lin, Dai, Zhou, Su, Chen, Qiao, Wang, and Luo}]{Wang2025InternVL35AO}
Weiyun Wang, Zhangwei Gao, Lixin Gu, Hengjun Pu, Long Cui, Xingguang Wei, Zhaoyang Liu, Linglin Jing, Shenglong Ye, Jie Shao, Zhaokai Wang, Zhe Chen, Hongjie Zhang, Ganlin Yang, Haomin Wang, Qi~Wei, Jinhui Yin, Wenhao Li, Erfei Cui, and 44 others. 2025{\natexlab{b}}.
\newblock \href {https://api.semanticscholar.org/CorpusID:280710824} {Internvl3.5: Advancing open-source multimodal models in versatility, reasoning, and efficiency}.
\newblock \emph{ArXiv}, abs/2508.18265.

\bibitem[{Wei et~al.(2022)Wei, Wang, Schuurmans, Bosma, Chi, Xia, Le, and Zhou}]{Wei2022ChainOT}
Jason Wei, Xuezhi Wang, Dale Schuurmans, Maarten Bosma, Ed~H. Chi, F.~Xia, Quoc Le, and Denny Zhou. 2022.
\newblock \href {https://api.semanticscholar.org/CorpusID:246411621} {Chain of thought prompting elicits reasoning in large language models}.
\newblock \emph{ArXiv}, abs/2201.11903.

\bibitem[{Wu et~al.(2024)Wu, Zhao, Saxon, Bui, Wang, Zhang, and Chang}]{Wu2024VSPAT}
Qiucheng Wu, Handong Zhao, Michael~Stephen Saxon, Trung~M. Bui, William~Yang Wang, Yang Zhang, and Shiyu Chang. 2024.
\newblock \href {https://api.semanticscholar.org/CorpusID:270878452} {Vsp: Assessing the dual challenges of perception and reasoning in spatial planning tasks for vlms}.
\newblock \emph{ArXiv}, abs/2407.01863.

\bibitem[{Xu et~al.(2025)Xu, Wang, Wang, Chen, gang Zhou, Yang, Lu, Li, Wang, Zhu, Wang, Dai, and Zhu}]{Xu2025VisuLogicAB}
Weiye Xu, Jiahao Wang, Weiyun Wang, Zhe Chen, Wen gang Zhou, Aijun Yang, Lewei Lu, Houqiang Li, Xiaohua Wang, Xizhou Zhu, Wenhai Wang, Jifeng Dai, and Jinguo Zhu. 2025.
\newblock \href {https://api.semanticscholar.org/CorpusID:277954881} {Visulogic: A benchmark for evaluating visual reasoning in multi-modal large language models}.
\newblock \emph{ArXiv}, abs/2504.15279.

\bibitem[{Yang et~al.(2025)Yang, Yu, Chen, Shen, and Gan}]{Yang2025MachineMI}
Zeyuan Yang, Xueyang Yu, Delin Chen, Maohao Shen, and Chuang Gan. 2025.
\newblock \href {https://api.semanticscholar.org/CorpusID:279464966} {Machine mental imagery: Empower multimodal reasoning with latent visual tokens}.
\newblock \emph{ArXiv}, abs/2506.17218.

\bibitem[{Zhang et~al.(2025{\natexlab{a}})Zhang, Zhong, Xia, Yu, Li, He, Shu, Liu, She, Wang, and Jiang}]{Zhang2025CMMCoTEC}
Guanghao Zhang, Tao Zhong, Yan Xia, Zhelun Yu, Haoyuan Li, Wanggui He, Fangxun Shu, Mushui Liu, Dong She, Yi~Wang, and Hao Jiang. 2025{\natexlab{a}}.
\newblock \href {https://api.semanticscholar.org/CorpusID:276884562} {Cmmcot: Enhancing complex multi-image comprehension via multi-modal chain-of-thought and memory augmentation}.
\newblock \emph{ArXiv}, abs/2503.05255.

\bibitem[{Zhang et~al.(2025{\natexlab{b}})Zhang, Wu, Li, Shang, Xia, Huang, Zhang, Dong, Zhang, Wang, Tan, and Wei}]{Zhang2025LatentSS}
Huanyu Zhang, Wenshan Wu, Chengzu Li, Ning Shang, Yan Xia, Yangyu Huang, Yifan Zhang, Li~Dong, Zhang Zhang, Liang Wang, Tien-Ping Tan, and Furu Wei. 2025{\natexlab{b}}.
\newblock \href {https://api.semanticscholar.org/CorpusID:282400662} {Latent sketchpad: Sketching visual thoughts to elicit multimodal reasoning in mllms}.
\newblock \emph{ArXiv}, abs/2510.24514.

\bibitem[{Zhang et~al.(2023)Zhang, Zhang, Li, Zhao, Karypis, and Smola}]{Zhang2023MultimodalCR}
Zhuosheng Zhang, Aston Zhang, Mu~Li, Hai Zhao, George Karypis, and Alexander~J. Smola. 2023.
\newblock \href {https://api.semanticscholar.org/CorpusID:256504063} {Multimodal chain-of-thought reasoning in language models}.
\newblock \emph{Trans. Mach. Learn. Res.}, 2024.

\end{thebibliography}
\clearpage
\appendix

\setcounter{table}{4} 
\setcounter{figure}{5} 
\setcounter{page}{1}

\section{Additional Experimental Results}
\label{app:additional_results}

In this section, we provide supplementary comparisons to further validate the effectiveness of the Interleaved Latent Visual Reasoning (ILVR) framework.

\subsection{Comparison with Latent Visual Reasoning (LVR)}
As discussed in the main paper, a direct comparison with the original LVR framework~\citep{Li2025LatentVR} on the Zebra-CoT~\citep{Li2025ZebraCoTAD} dataset is not feasible because LVR relies on ground-truth bounding box (BBox) annotations.
To ensure a rigorous comparison, we adopted the LVR experimental protocol by training both the LVR baseline and our ILVR model on the Visual-CoT~\citep{Shao2024VisualCA} dataset (80k samples). 

The results in Table~\ref{tab:lvr_results} show that ILVR achieves superior generalization, particularly on VisuLogic~\citep{Xu2025VisuLogicAB} (+4.5\% average accuracy) and Zebra-CoT (+0.3\% average accuracy). We attribute this improvement to the nature of feature selection. While LVR relies on bounding box annotations to strictly localize regions deemed important by humans, such explicit supervision may not always align with the intrinsic features required by the model for reasoning. In contrast, our momentum teacher autonomously selects visual features based on the current reasoning context. This suggests that adaptively distilled features, which are optimized for the model's own latent space, provide more effective guidance than rigid human-defined regions, thereby leading to better performance on unseen tasks.
\begin{table}[h]
    \centering
    \caption{\textbf{Comparison with LVR fine-tuned on Visual-CoT.} Models are evaluated on OOD benchmarks.}
    \label{tab:lvr_results}
    \small
    \setlength{\tabcolsep}{2pt}
    \resizebox{\columnwidth}{!}{
    \begin{tabular}{ll ccc c}
    \toprule
    \textbf{Model} & \textbf{Paradigm} & \textbf{EMMA} & \textbf{VisLog} & \textbf{Zebra} & \textbf{Total} \\
    \midrule
    LVR & Direct & \textbf{24.0\%} & 22.1\% & 47.0\% & 31.9\% \\
    \rowcolor{ourblue} \textbf{ILVR (Ours)} & \textbf{Interleaved} & 21.5\% & \textbf{26.6\%} & \textbf{47.3\%} & \textbf{32.6\%} \\
    \bottomrule
    \end{tabular}
    }
\end{table}

\subsection{Comparison with Sketchpad}
\label{sec:sketchpad_comparison}
We further compare ILVR against Sketchpad~\citep{Zhang2025LatentSS}. Before analyzing the results, it is important to note a disclaimer regarding the reproduction of the Sketchpad baseline. We encountered data processing discrepancies in the official repository, which prevented direct execution. We have resolved these issues to the best of our ability to establish a functional baseline; however, these results should be considered tentative and may be updated pending future fixes to the official implementation.

Table~\ref{tab:sketchpad_results} details the performance on OOD benchmarks for models fine-tuned on the Zebra-CoT (10k) subset. ILVR achieves a Total Average accuracy of 37.5\%, significantly outperforming Sketchpad's 33.0\%. We attribute this performance advantage to the superior efficacy of our teacher-guided feature selection over Sketchpad's alignment mechanism. Sketchpad operates by projecting hidden states into the vision encoder's space (prior to LLM projection), forcing them to align with 256 visual tokens derived from a resized $448 \times 448$ helper image, and subsequently projecting them back into the LLM space to aid reasoning. This process essentially enforces a rigid alignment with the overall features of the helper image. In contrast, ILVR employs a momentum teacher to actively select features. Instead of aligning to the entire feature map, our teacher dynamically identifies and distills the specific visual cues that are most beneficial for the current reasoning context. This selective mechanism provides more precise and effective guidance than Sketchpad's global alignment strategy.

\begin{table}[h]
    \centering
    \caption{\textbf{Comparison with Sketchpad.} ILVR results correspond to Stage 2.}
    \label{tab:sketchpad_results}
    \small
    \setlength{\tabcolsep}{2pt}
    \resizebox{\columnwidth}{!}{
    \begin{tabular}{ll ccc c}
    \toprule
    \textbf{Model} & \textbf{Paradigm} & \textbf{EMMA} & \textbf{VisLog} & \textbf{Zebra} & \textbf{Total} \\
    \midrule
    Sketchpad & Direct & 25.0\% & 25.9\% & 43.8\% & 33.0\% \\
    \rowcolor{ourblue} \textbf{ILVR (Ours)} & \textbf{Interleaved} & \textbf{33.3\%} & \textbf{29.3\%} & \textbf{47.8\%} & \textbf{37.5\%} \\ 
    \bottomrule
    \end{tabular}
    }
\end{table}

\section{Implementation Details}
\label{app:implementation}

\subsection{Training Infrastructure \& Setup}
All models were trained on a cluster of $8 \times$ NVIDIA H200 GPUs using DeepSpeed Zero-3 optimization with \texttt{Qwen2.5-VL-7B} backbone. 
We use the AdamW optimizer with a cosine learning rate scheduler. To prevent overfitting on the limited 10k Zebra-CoT subset, we apply a weight decay of 0.01 and a moderate warmup ratio. The specific hyperparameters for each stage are detailed in Table~\ref{tab:hyperparameters}.

\begin{table}[h]
    \centering
    \caption{\textbf{Hyperparameters for ILVR Training.}}
    \label{tab:hyperparameters}
    \resizebox{\columnwidth}{!}{%
    \begin{tabular}{lcc}
        \toprule 
        Hyperparameter & Stage 1 & Stage 2 \\
        \midrule 
        Learning Rate & $1 \times 10^{-5}$ & $1 \times 10^{-5}$ \\
        Batch Size & 1 & 1 \\
        Gradient Accumulation & 8 & 1 \\
        Latent Tokens ($K$) & 8 & 8 \\
        Align Weight ($\lambda_{\text{sim}}$) & 1.0 & N/A \\
        EMA Decay ($\tau$) & 0.999 & N/A \\
        Epochs & 15 / 2 & 15 / 1.5 \\
        \bottomrule
    \end{tabular}%
    }
\end{table}

\subsection{Data Construction Pipeline}
\label{app:data_construction}
We construct the training data to support the interleaved text-latent paradigm. Each data sample is formatted as a conversation containing a user query and a multi-step assistant response.

Below is a simplified example of the data format using a chat template structure. We highlight the interleaved nature of the assistant's response:

\newcommand{\jsonkey}[1]{\textcolor{blue!60!black}{\textbf{"#1"}}}
\newcommand{\jsonval}[1]{"#1"}
\newcommand{\jsoncomment}[1]{\textcolor{gray}{\textit{#1}}}

\begin{tcolorbox}[colback=gray!5, colframe=gray!60, arc=2mm, title=\textbf{Data Sample Format (Chat Template)}]
\small\ttfamily 
[ \\
\hspace*{1em} \{ \\
\hspace*{2em} \jsonkey{role}: \jsonval{user}, \\
\hspace*{2em} \jsonkey{content}: [ \\
\hspace*{3em} \{ \jsonkey{type}: \jsonval{image}, \jsonkey{image}: \jsonval{original\_input.jpg} \}, \\
\hspace*{3em} \{ \jsonkey{type}: \jsonval{text}, \jsonkey{text}: \jsonval{How many red objects are to the left...?} \} \\
\hspace*{2em} ] \\
\hspace*{1em} \}, \\
\hspace*{1em} \{ \\
\hspace*{2em} \jsonkey{role}: \jsonval{assistant}, \\
\hspace*{2em} \jsonkey{content}: [ \\
\hspace*{3em} \{ \jsonkey{type}: \jsonval{text}, \jsonkey{text}: \jsonval{First, I need to locate the blue cube...} \}, \\
\hspace*{3em} \{ \jsonkey{type}: \jsonval{image}, \jsonkey{image}: \jsonval{crop\_blue\_cube.jpg} \}, \jsoncomment{\% Becomes <latent>} \\
\hspace*{3em} \{ \jsonkey{type}: \jsonval{text}, \jsonkey{text}: \jsonval{Now I will scan the area to its left...} \}, \\
\hspace*{3em} \{ \jsonkey{type}: \jsonval{image}, \jsonkey{image}: \jsonval{left\_region\_red\_filter.jpg} \}, \jsoncomment{\% Becomes <latent>} \\
\hspace*{3em} \{ \jsonkey{type}: \jsonval{text}, \jsonkey{text}: \jsonval{I see two red spheres. The answer is 2.} \} \\
\hspace*{2em} ] \\
\hspace*{1em} \} \\
]
\end{tcolorbox}

\section{Detailed Dataset Composition}
\label{app:dataset_details}

To evaluate the robustness and versatility of our framework, we curate a diverse suite of benchmarks encompassing both in-distribution (IID) and out-of-distribution (OOD) settings. 

For in-distribution evaluation, we focus on fine-grained visual perception and sequential planning using the COMT and VSP datasets. To further assess generalizability, we construct a strictly controlled subset from the Zebra-CoT dataset as our OOD benchmark, challenging the model with multi-step reasoning across scientific and logic domains. Table~\ref{tab:dataset_stats} provides a comprehensive summary of the statistics and task definitions.

\begin{table}[h]
    \centering
    \caption{\textbf{Summary of Dataset Composition.} Key characteristics for the three primary benchmarks.}
    \label{tab:dataset_stats}
    \small
    \setlength{\tabcolsep}{3pt}
    \renewcommand{\arraystretch}{1.1}
    \resizebox{\columnwidth}{!}{
    \begin{tabular}{l c p{4.5cm}}
    \toprule
    \textbf{Dataset} & \textbf{Split} & \textbf{Key Characteristics} \\
    \midrule
    \rowcolor{gray!10} 
    \textbf{COMT} & \begin{tabular}{@{}c@{}}3.4k / 400\\(IID)\end{tabular} & \textbf{Atomic Manipulation}: Fine-grained perception tasks (Creation, Deletion, Selection, Update). \\
    
    \textbf{VSP} & \begin{tabular}{@{}c@{}}1k / 400\\(IID)\end{tabular} & \textbf{Sequential Planning}: Tracking visual state changes over long horizons. \\
    
    \rowcolor{gray!10} 
    \textbf{Zebra-CoT} & \begin{tabular}{@{}c@{}}10k / -\\(OOD)\end{tabular} & \textbf{Complex Reasoning}: \newline 1. \textit{Science}: Physics, Graphs. \newline 2. \textit{Logic}: Chess, Ciphers, Maze, Tetris, RPM \newline 3. \textit{3D}: Counting, Planning, Embodied. \\
    \bottomrule
    \end{tabular}
    }
\end{table}
\subsection{Case Studies}

To illustrate the generalization capabilities of ILVR, we present a series of inference cases using the Qwen2.5-VL-7B backbone. For visualization purposes, we represent the latent reasoning segments with <|latent\_start|><|latent\_pad|><|latent\_end|>.

\begin{figure*}[p]
    \centering
    \vfill
    \includegraphics[width=\textwidth,height=0.92\textheight,keepaspectratio]{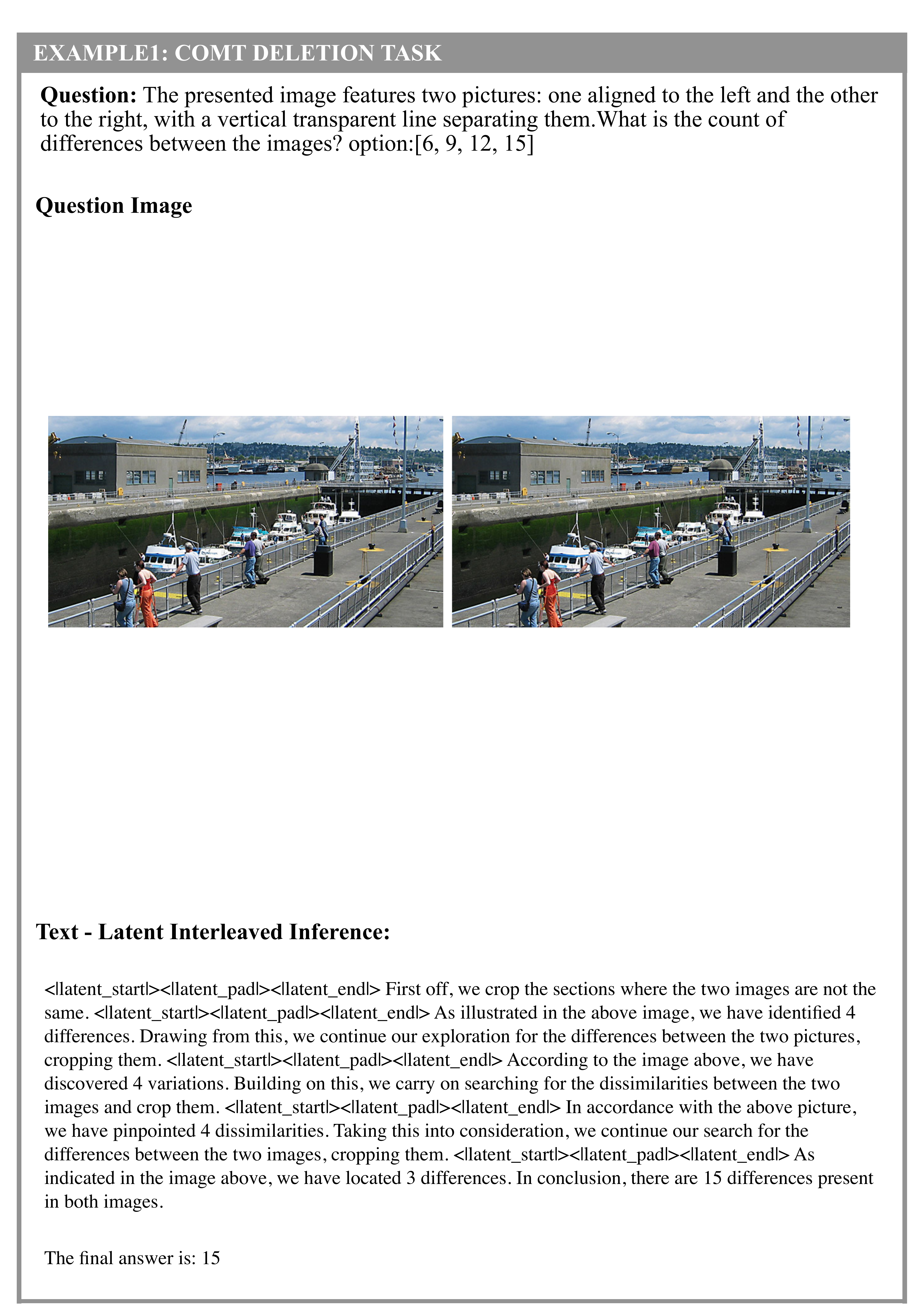}
    \vfill
    \caption{Example of COMT Deletion Task}
    \label{fig:case1}
\end{figure*}

\begin{figure*}[p]
    \centering
    \vfill
    \includegraphics[width=\textwidth,height=0.92\textheight,keepaspectratio]{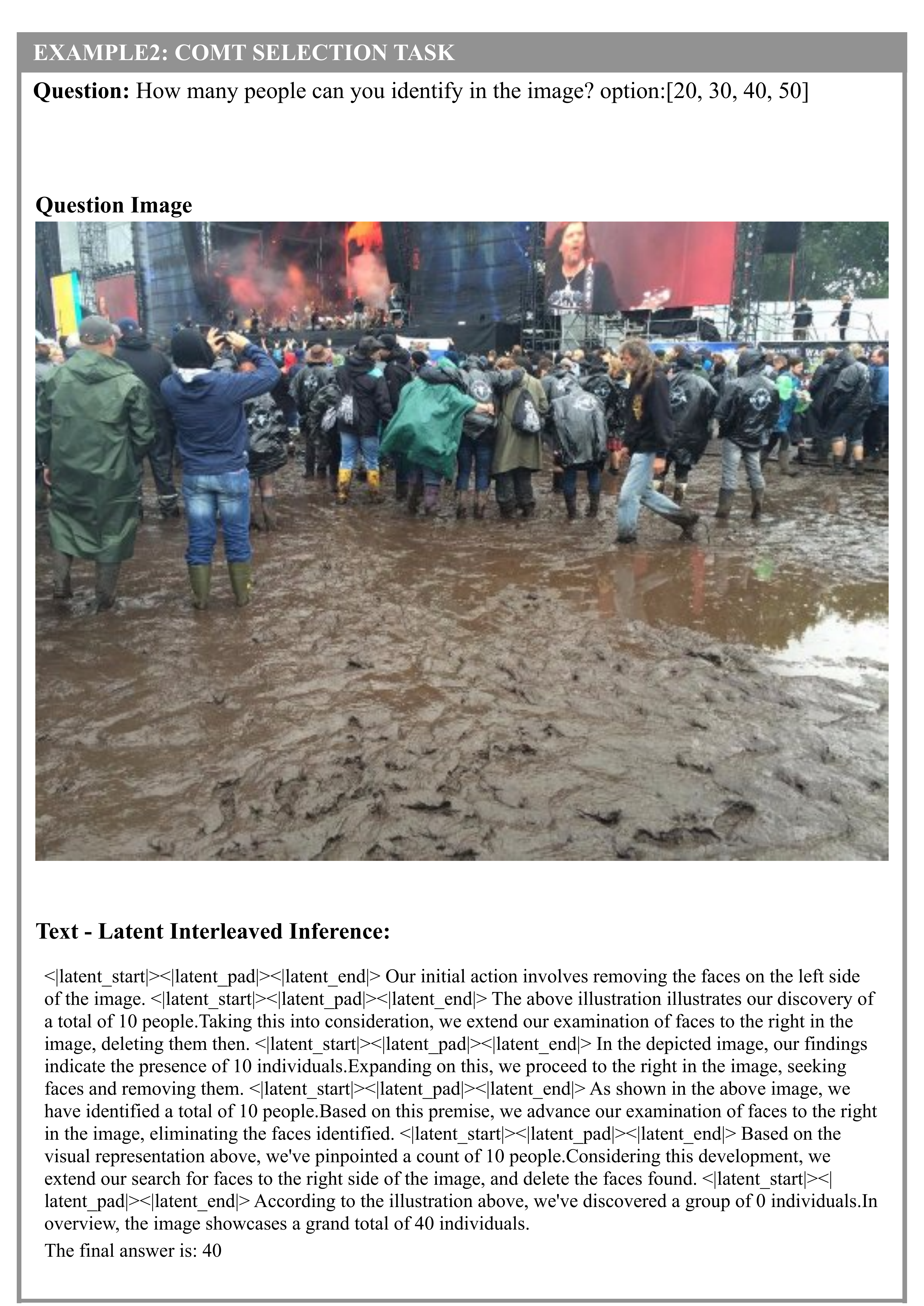}
    \vfill
    \caption{Example of COMT Selection Task}
    \label{fig:case2}
\end{figure*}

\begin{figure*}[p]
    \centering
    \vfill
    \includegraphics[width=\textwidth,height=0.92\textheight,keepaspectratio]{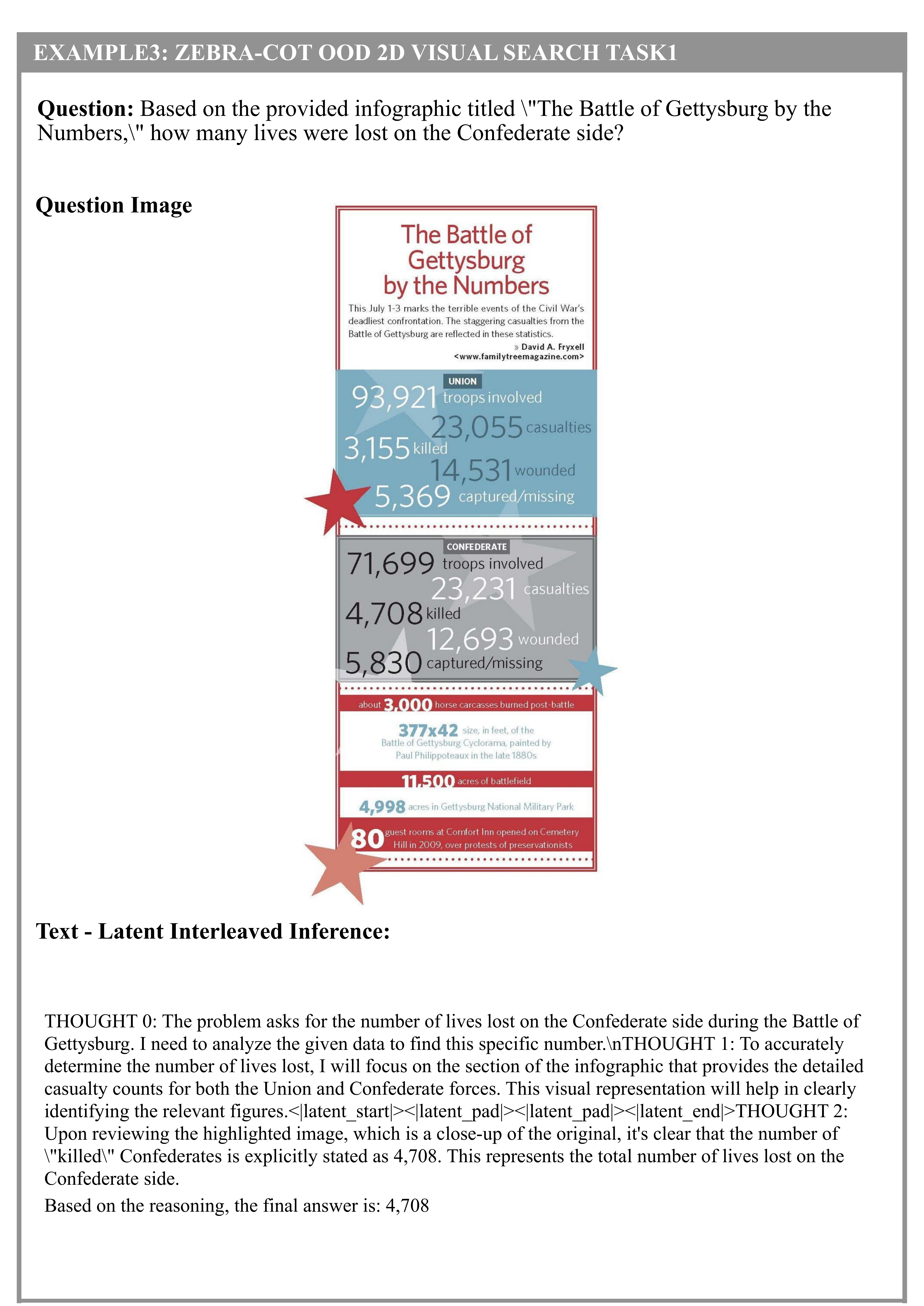}
    \vfill
    \caption{Example of Zebra-CoT 2D Visual Search (Task 1)}
    \label{fig:case3}
\end{figure*}

\begin{figure*}[p]
    \centering
    \vfill
    \includegraphics[width=\textwidth,height=0.92\textheight,keepaspectratio]{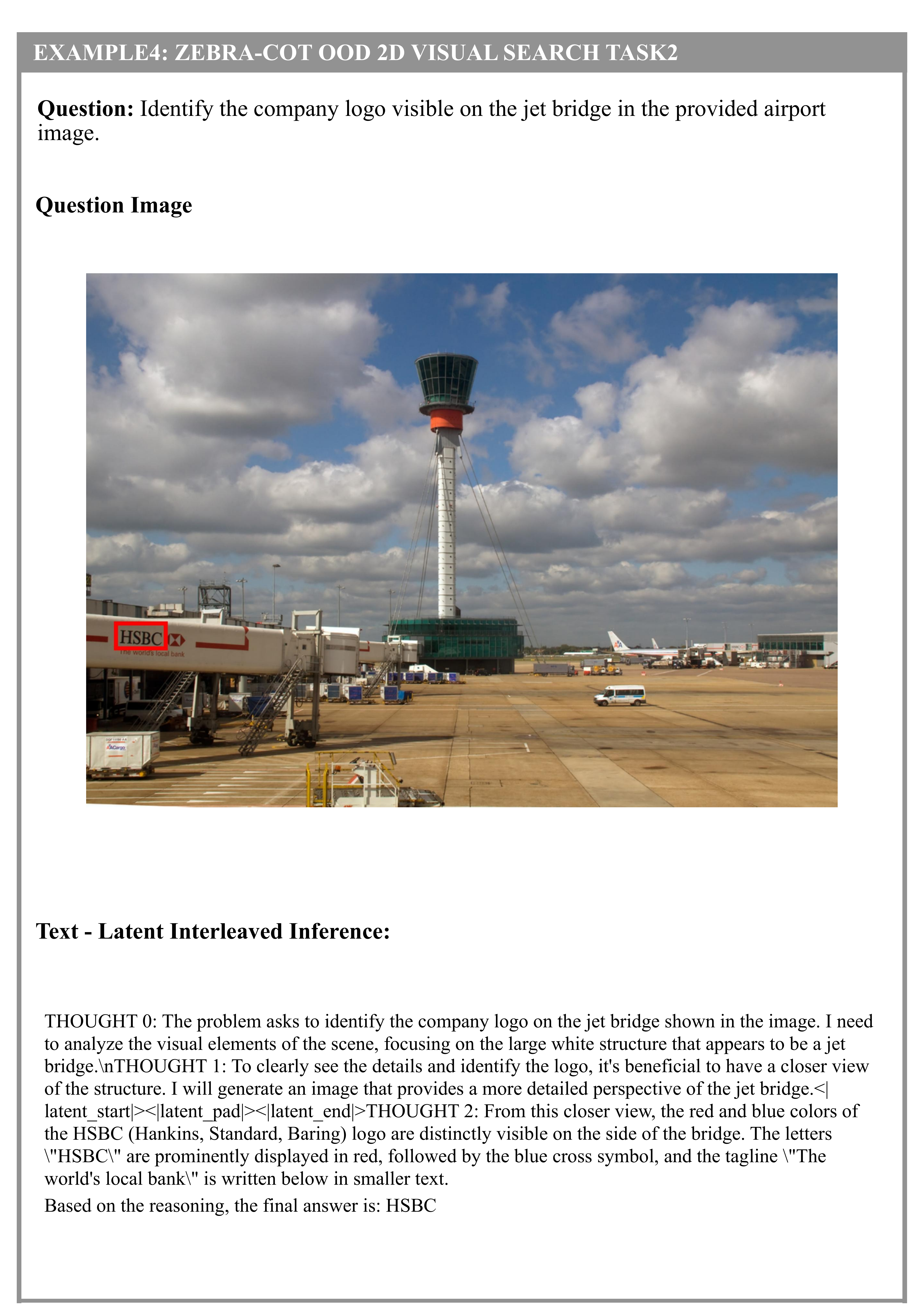}
    \vfill
    \caption{Example of Zebra-CoT 2D Visual Search (Task 2)}
    \label{fig:case4}
\end{figure*}

\begin{figure*}[p]
    \centering
    \vfill
    \includegraphics[width=\textwidth,height=0.92\textheight,keepaspectratio]{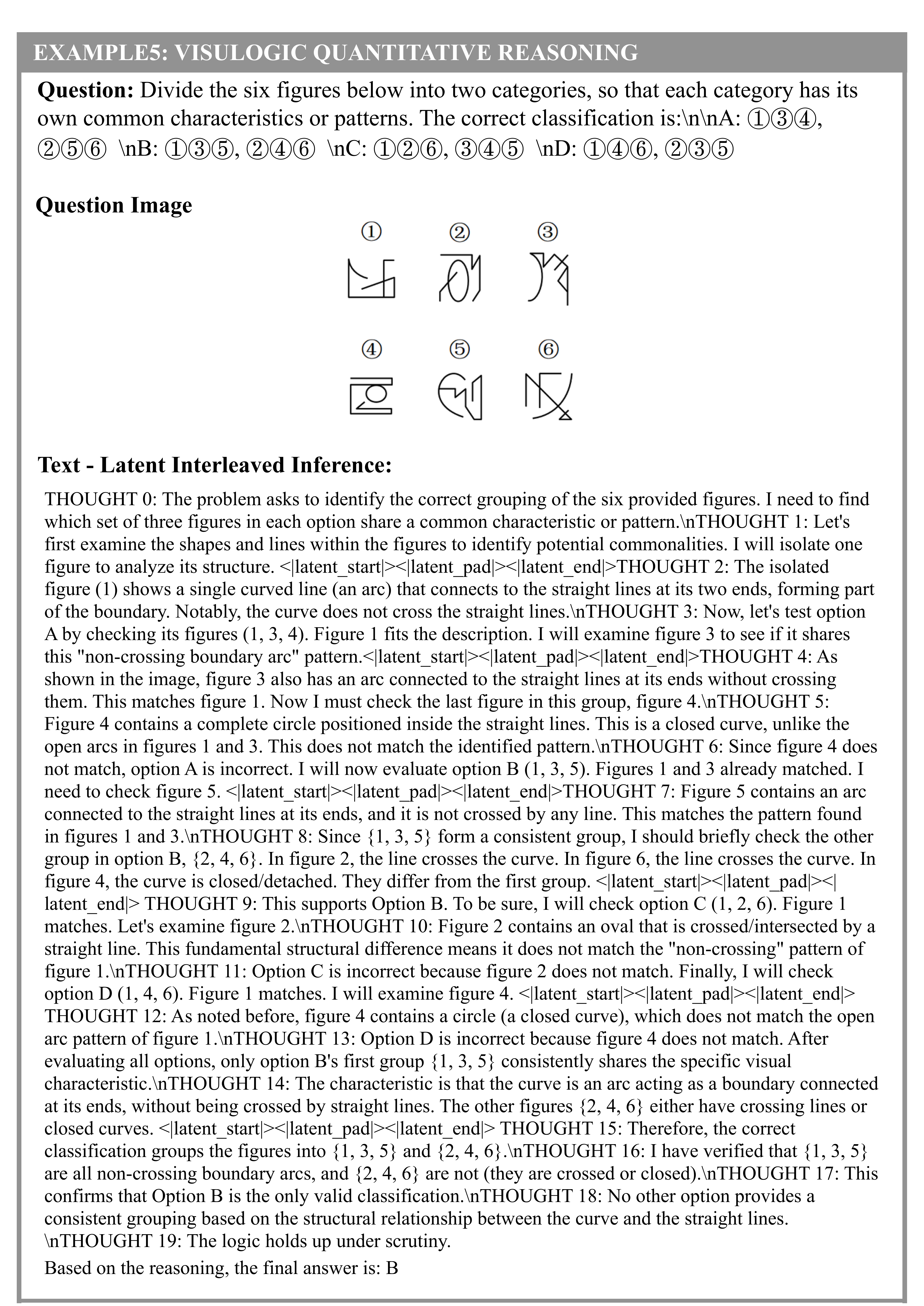}
    \vfill
    \caption{Example of VisuLogic Quantitative Reasoning }
    \label{fig:case5}
\end{figure*}

\begin{figure*}[p]
    \centering
    \vfill
    \includegraphics[width=\textwidth,height=0.92\textheight,keepaspectratio]{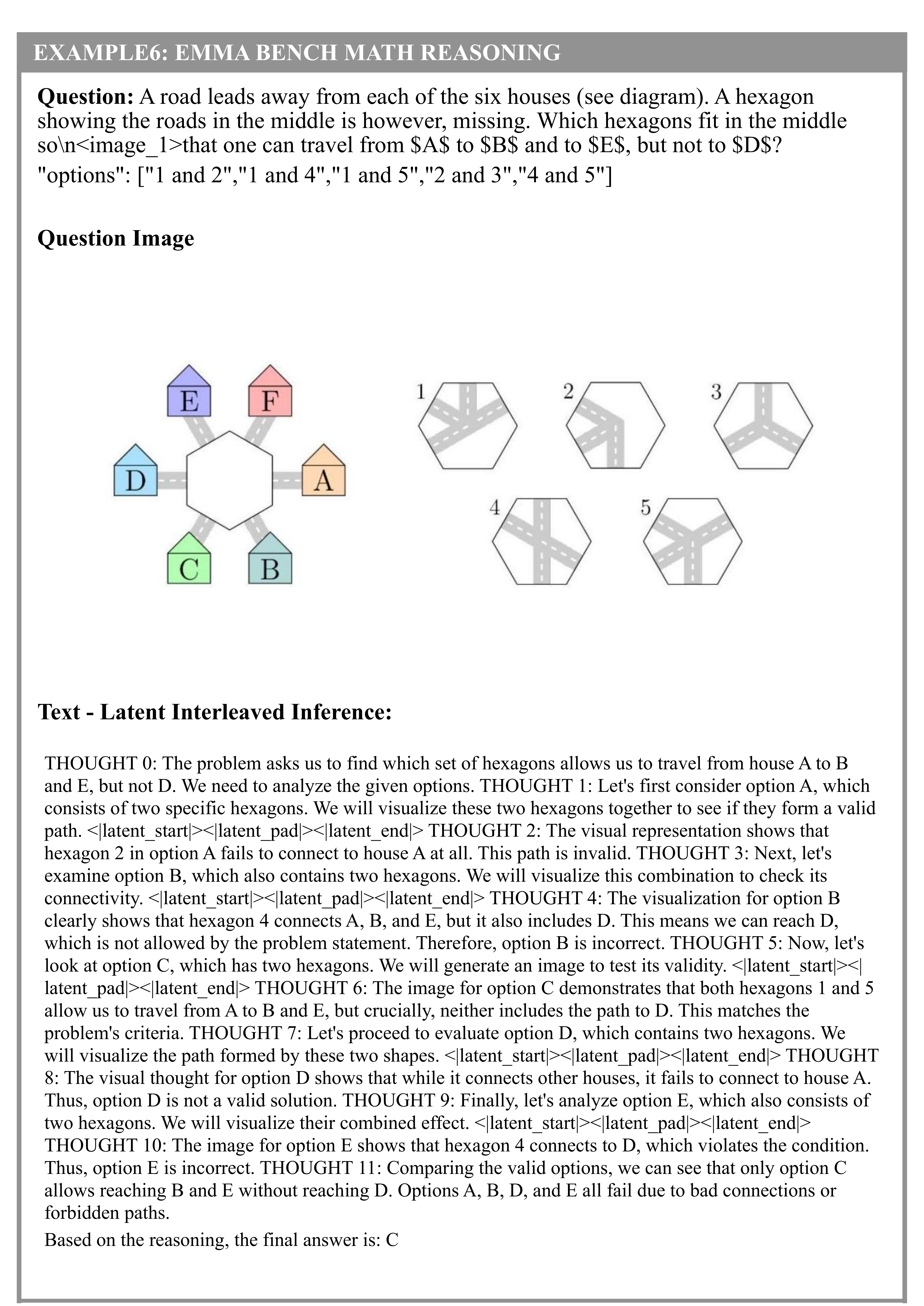}
    \vfill
    \caption{Example of EMMA Bench Math Reasoning }
    \label{fig:case6}
\end{figure*}

The tasks of the selected examples include fine-grained perception on COMT (Selection and Deletion, Fig.~\ref{fig:case1} and Fig.~\ref{fig:case2}) , 2D visual search on held-out Zebra-CoT (Fig.~\ref{fig:case3} and Fig.~\ref{fig:case4}) , visual logic reasoning on VisuLogic (Quantitative Reasoning, Fig.~\ref{fig:case5}) , and Math reasoning on EMMA (Fig.~\ref{fig:case6}).
Note that ILVR effectively utilizes latent thinking to model evolving states. For example, in the Deletion and Position tasks, the generated latent representations dynamically update to reflect the removal of objects or the simulation of movement trajectories, rather than relying on static visual features.
\newpage

\clearpage

\clearpage

\clearpage

\clearpage

\clearpage

\clearpage

\end{document}